\documentclass[10pt,twocolumn,letterpaper]{article}
\ifdefined\pdfsuppressptexinfo\pdfsuppressptexinfo=-1\fi

\usepackage[pagenumbers]{cvpr}

\renewcommand{\paragraph}[1]{\vspace{.5em}\noindent\textbf{#1}\ }

\usepackage[T1]{fontenc}
\usepackage{amsthm}
\usepackage{multirow}
\usepackage{colortbl}

\newcommand{\method}{FedSO\xspace}            %

\definecolor{oursgray}{gray}{0.93}
\newcommand{\oursrow}{\rowcolor{oursgray}}

\newcommand{\ty}{\tilde{y}}
\newcommand{\tY}{\tilde{Y}}
\newcommand{\E}{\mathbb{E}}
\newcommand{\Prob}{\mathbb{P}}
\newcommand{\ind}{\mathbf{1}}
\newcommand{\wstar}{w^{\star}}
\newcommand{\hw}{\hat{w}}
\newcommand{\Rcor}{R_{\mathrm{cor}}}

\newtheorem{proposition}{Proposition}
\newtheorem{corollary}{Corollary}

\theoremstyle{remark}

\AtEndPreamble{%
  \crefname{proposition}{Prop.}{Props.}
  \crefname{corollary}{Cor.}{Cors.}
  \crefname{remark}{Remark}{Remarks}
  \crefname{lemma}{Lemma}{Lemmas}
}
\makeatletter
\@ifpackageloaded{lineno}{%
  \ifdefined\IfFormatAtLeastTF
    \IfFormatAtLeastTF{2025-06-01}{%
      \RemoveFromHook{build/column/before}[lineno]%
      \AddToHook{build/column/after}[fedso]{%
        \setbox\@outputbox\vbox{\boxmaxdepth\@maxdepth
          \protected@write\@auxout{}{\string\def\string\@LN@column{\if@firstcolumn1\else2\fi}}%
          \box\@outputbox}}%
    }{}%
  \fi
}{}
\makeatother

\definecolor{cvprblue}{rgb}{0.21,0.49,0.74}
\usepackage[pagebackref,breaklinks,colorlinks,allcolors=cvprblue]{hyperref}

\title{A Second Opinion for Every Label:\\Federated Learning under Client-Specific Label Noise}

\author{%
Jiankang Peng$^{1*}$\quad Bixiao Zeng$^{1*\dagger}$\quad Yiqiang Chen$^{1\dagger}$\quad Xiaoke Deng$^{1}$\\
Hanlin Gu$^{2}$\quad Xu Guo$^{3}$\quad Yingwei Zhang$^{1}$\quad Yang Gu$^{1}$\\[0.4em]
$^{1}$Institute of Computing Technology, Chinese Academy of Sciences\\
$^{2}$The Hong Kong University of Science and Technology\\
$^{3}$LILY Research Centre, Nanyang Technological University, Singapore
}

\begin{document}
\maketitle
\begin{NoHyper}{\makeatletter\renewcommand{\@makefntext}[1]{\noindent #1}\makeatother\footnotetext{$^{*}$Equal contribution.\quad $^{\dagger}$Corresponding authors.\\{\tt \{pengjiankang26z, zengbixiao, yqchen\}@ict.ac.cn}}}\end{NoHyper}
\begin{abstract}
In federated learning with noisy labels, clients differ in class composition and labeling errors, so the same prediction and observed label can warrant different trust across clients. Existing methods typically assess label reliability using models trained on the labels they judge, but these in-sample signals can reflect memorization rather than correctness. We propose \textbf{Fed}erated \textbf{S}econd \textbf{O}pinion (\method), which combines out-of-fold predictions with each client's labeling signature to estimate the posterior probability that an observed label is correct. The signature pairs the client's class prior and noise transition matrix to infer trust from a prediction by Bayes' rule. Because the signature scales the predictive odds, the prediction source decides what this inference does, restoring trust in ambiguous but clean labels on out-of-fold predictions while reinforcing memorized errors on in-sample ones. Cross-fitted federated teachers provide these second opinions without training on the labels they score. Each client infers frozen trust weights through its fitted signature, and a federated student trains on all original labels with these weights. On a synthetic noise spectrum from concentrated to dispersed flips (CIFAR-10/100, intracranial hemorrhage CT) and on real-world noise (CIFAR-10N/100N), \method outperforms the strongest of fifteen baselines by 2.8 to 6.2 points and stays ahead from 10 to 100 clients.
\end{abstract}

\section{Introduction}
\label{sec:intro}

\begin{figure}[t]
  \centering
  \includegraphics[width=\linewidth]{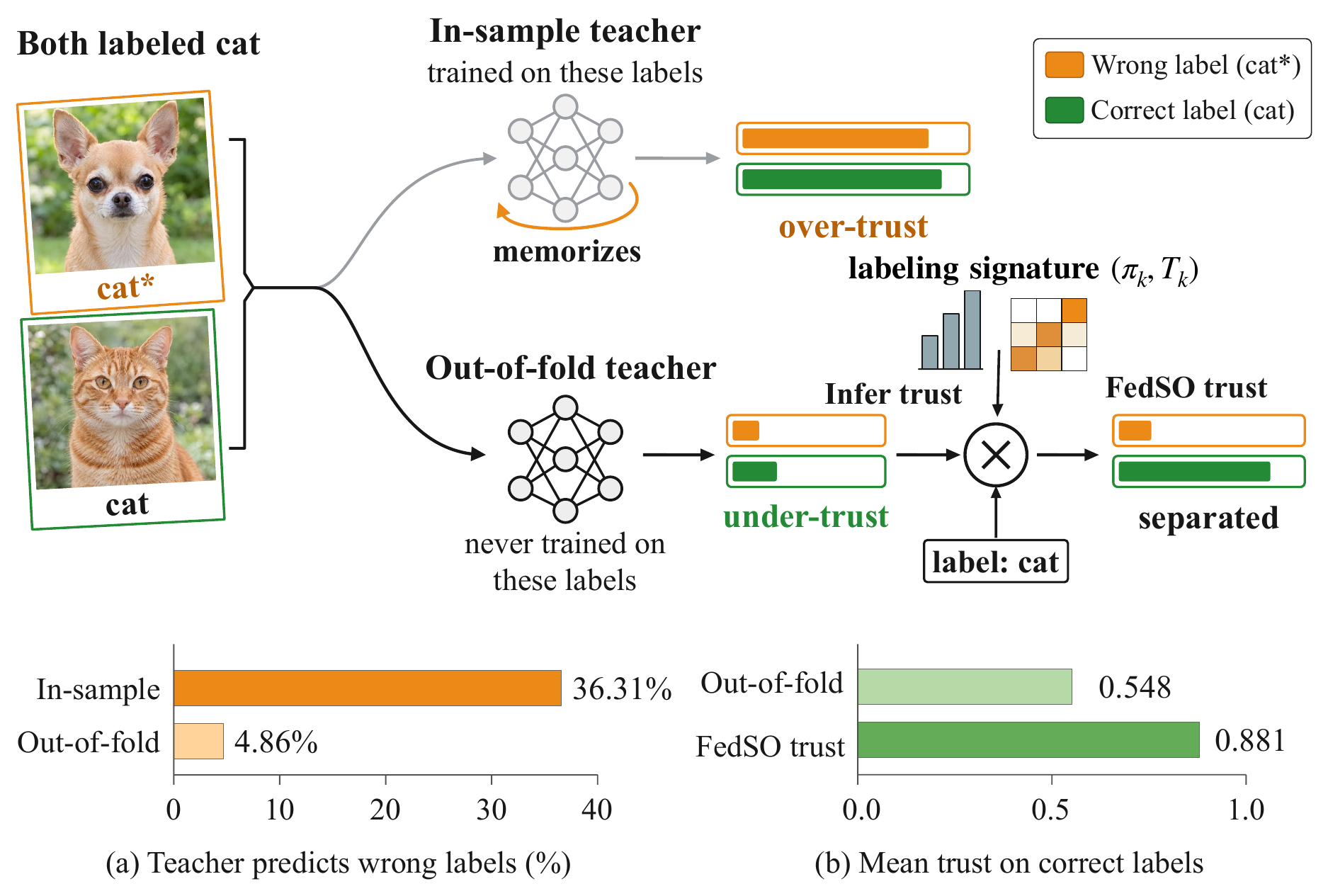}
  \caption{\textbf{Three verdicts on the same labels.}
  Two images that a client labeled ``cat'', one wrongly (cat*); bar lengths are illustrative trust scores in $[0,1]$.
  \emph{Top:} a teacher trained on these labels memorizes them and trusts both (over-trust).
  \emph{Middle, left:} used alone, the score of a teacher that never trained on them doubts both (under-trust).
  \emph{Middle, right:} \method infers trust from this score through the client's labeling signature, its class prior and transition matrix, which separates the two.
  \emph{Bottom:} (a)~share of wrong labels the teacher predicts and (b)~mean trust on correctly labeled samples, over the CIFAR-10 training labels scored by both teachers; on CIFAR-100 they are 59.2\% \vs 0.6\% and 0.32 \vs 0.85.}
  \label{fig:teaser}
\end{figure}

Federated learning (FL) trains a shared model while images and labels stay on the clients~\cite{fedavg}.
Each site annotates its own data with nobody to cross-check it, so clients differ not only in which classes they hold but also in how they err.
One client keeps confusing a particular pair of classes, while another slips across many.
Under such \emph{client-specific label noise}, a model's confidence does not measure how reliable a label is, and the same prediction and observed label can deserve very different trust at different clients.
An ambiguous image labeled ``cat'' is far less convincing from a client that often writes dogs as cats than from one that rarely does.
Turning a prediction into trust thus needs the client's \emph{labeling signature}, its class prior together with its noise transition matrix.

Existing methods for federated noisy-label learning act at different levels.
Client-level methods reweight or adjust noisy clients~\cite{fedcorr,fednoro,fedned}, sample-level methods filter or relabel examples by their losses and predictions~\cite{fedrn,fedelc,feddiv}, and others model client noise with transition matrices~\cite{fedbeat,fedefc} or representation geometry~\cite{fedrg}.
Their signals all come from a model that is trained on the very labels it judges.
Deep networks eventually memorize wrong labels~\cite{arpit2017closer,zhang2017rethinking}, so a confident prediction for the observed class may reflect memorization rather than the image (\cref{fig:teaser}, top).
A labeling signature cannot repair such a prediction.
Bayes' rule turns a prediction into trust by multiplying its odds with a client factor built from the signature, and this factor does not know where the odds came from.
Wherever it exceeds one, it makes memorized wrong labels look even more reliable.

Scoring each sample with a model trained on other data, an out-of-fold model, removes this path, because a model never trained on a label cannot have memorized it.
Yet its score alone is not enough.
An out-of-fold prediction is evidence about the image, not yet a verdict on the client's label.
Used directly as trust, it only says how much the image looks like each class, and it under-trusts hard or ambiguous images even when their labels are right (\cref{fig:teaser}, middle).
Both failures break the same rule.
Bayes' rule turns evidence about the image into trust through the client's signature, and only an out-of-fold prediction is evidence about the image alone.
Skipping the signature under-trusts clean labels, and feeding it in-sample predictions over-trusts memorized ones.
On CIFAR-100, inferring trust through the signature raises student accuracy from 61.41\% to 68.03\% on out-of-fold scores but lowers it from 64.96\% to 63.30\% on in-sample scores (\cref{tab:ablation}).
Whether modeling client-specific noise helps thus depends on where the prediction comes from.

We propose \textbf{Fed}erated \textbf{S}econd \textbf{O}pinion (\method), which follows this rule within standard federated training (\cref{fig:method}).
Each client splits its data into two folds, and two teachers are trained with FedAvg, each on one fold of every client, so that each scores the other fold.
Aggregation lets the teachers learn what clients share, the appearance of each class, while clients that err in other directions dilute the flipping rule of any single one.
Each client then fits its labeling signature using these out-of-fold predictions and its observed labels, and infers a frozen trust for every sample, the posterior probability that its observed label is correct.
A global student finally trains with FedAvg on all original labels weighted by this trust.
\method keeps every label and changes only its weight.
Trust judges whether the observed label is right without naming a replacement, so it stays useful under dispersed noise, where the transition matrix gives no hint about which other class is true.

To test this across error patterns, we construct the noise spectrum, a synthetic benchmark family in which clients range from concentrated pair flips to dispersed symmetric flips (\cref{fig:spectrum}).
On three synthetic benchmarks (CIFAR-10/100 and intracranial hemorrhage CT) and two real-world ones (CIFAR-10N/100N), \method ranks first on every dataset and metric, ahead of the strongest of fifteen baselines by 2.8 to 6.2 points and from 10 to 100 clients.

\noindent\textbf{Contributions.}
\begin{itemize}[leftmargin=*,itemsep=1pt,topsep=2pt]
  \item \textbf{Federated label trust.} We estimate for every sample the posterior probability that its label is correct, inferred from the out-of-fold predictions of cross-fitted federated teachers through private client labeling signatures, and train the student on all original labels without pruning or relabeling.
  \item \textbf{The prediction source decides what trust inference does.} We prove that exact trust weighting recovers the risk on correctly labeled samples, and show through an odds decomposition and controlled ablations why the same inference rescues hard clean labels on out-of-fold predictions and reinforces memorized errors on in-sample ones.
  \item \textbf{The noise spectrum.} We design a synthetic benchmark from concentrated to dispersed client-specific flips and evaluate \method on five synthetic and real-world benchmarks, with diagnostics that trace label trust to student memorization.
\end{itemize}

\section{Related work}
\label{sec:related}

\paragraph{Learning with noisy labels.}
Robust losses bound the influence of any single label~\cite{gce,sce,agce}.
Sample selection keeps small-loss examples, often with two networks~\cite{coteaching,mentornet,dividemix}.
Loss correction models the noise with a transition matrix~\cite{natarajan2013learning,patrini2017making}.
Reweighting assigns each sample an importance, either from a density ratio~\cite{liu2016reweighting} or by meta-learning with clean validation supervision~\cite{ren2018learning,shu2019metaweight}.
Memorization of wrong labels sets in after an early phase of fitting clean ones~\cite{arpit2017closer,liu2020elr}.
Cross-validation methods for centralized training score each label with held-out predictions, estimate noise statistics, and prune or relabel the samples they flag~\cite{incv,confidentlearning}.
Self-supervised pretraining further makes training under label noise more robust~\cite{c2d,ghosh2021contrastive}.

\paragraph{Federated learning with noisy labels.}
Client-level methods identify noisy clients and treat them differently~\cite{fedcorr,fednoro,fedned}.
Sample-level methods select or relabel examples from the losses and predictions of local and global models~\cite{fedrn,fedelc,feddiv}.
Correction-based methods estimate transition matrices~\cite{fedbeat,fedefc}, and FedRG~\cite{fedrg} detects noisy labels from the geometry of a self-supervised representation that keeps training on the labels, with a personalized transition matrix per client.
Two-stage pipelines first pretrain without labels and then fine-tune robustly~\cite{fedltf,fedrg}.
Benchmarks for the area vary noise rates across clients and generate each client's noise within its local classes~\cite{fednoisy}. Our noise spectrum additionally varies the flip map and its spread between clients that share classes.
In each of these methods the evidence about a label comes from a model that trains on that label.
\method takes it from models that never trained on the label, and uses each client's labeling signature to infer trust rather than to correct the loss.

\paragraph{Annotator-specific noise.}
Modeling each annotator by a confusion matrix goes back to Dawid and Skene~\cite{dawidskene} and underlies learning from crowds~\cite{raykar2010crowds}.
Later work estimates annotator confusion matrices jointly with a classifier~\cite{khetan2018learning}.
In \method, out-of-fold federated predictions provide a reference beyond each client's labeling habits.
Each client uses this reference and its observed labels to fit its own class prior and transition matrix and infer trust.

\section{Problem setting}
\label{sec:setting}

\begin{figure}[t]
  \centering
  \includegraphics[width=\linewidth]{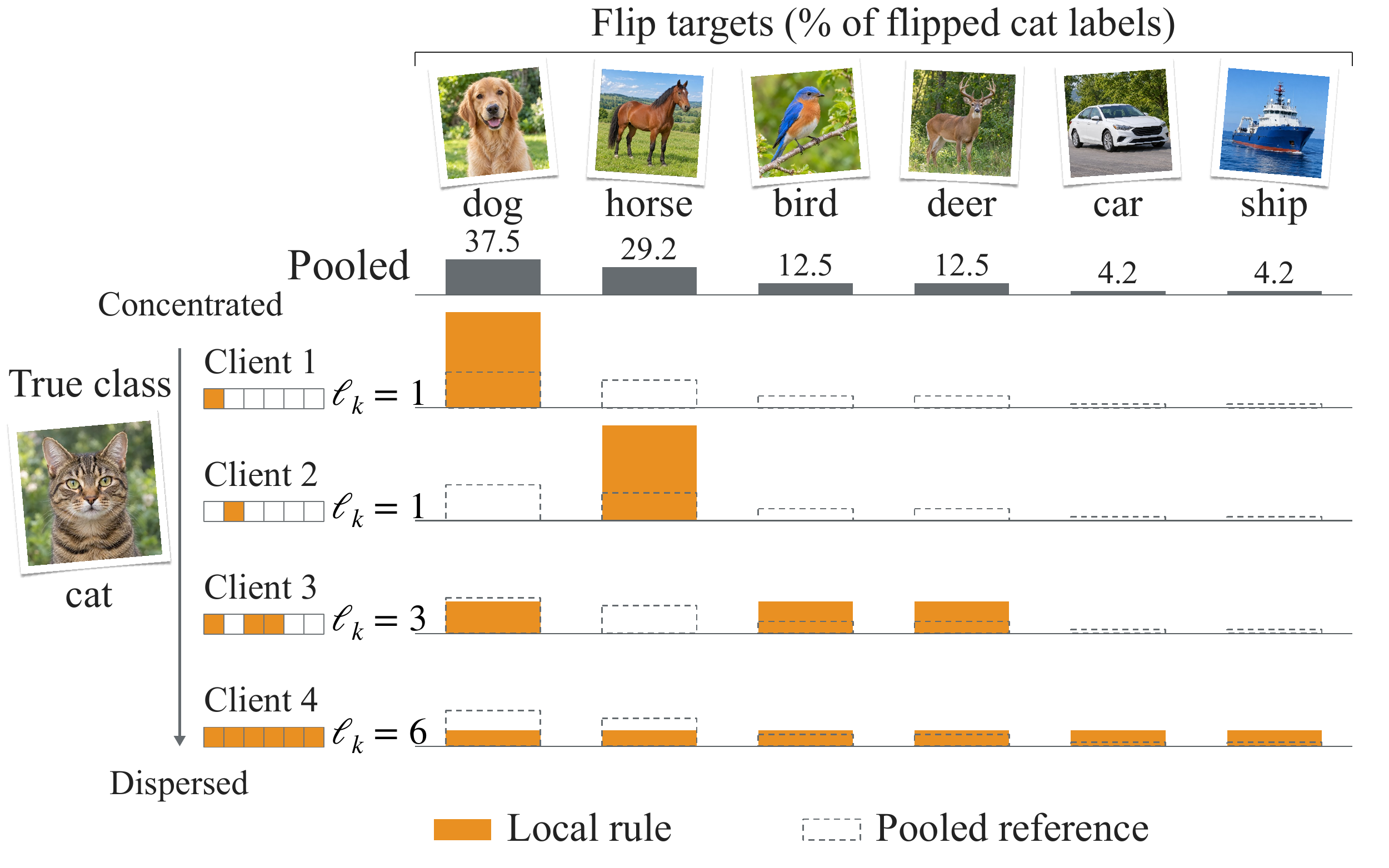}
  \caption{\textbf{The noise spectrum.}
  Images of one true class (here ``cat'') at four clients.
  Each client sends its flipped labels to its own targets: Client~1 to a single class, Client~4 to all others.
  Pooled over clients (top row), the flips spread over many classes and hide each client's rule.
  Filled: the client's rule; dashed: pooled reference.
  In our CIFAR-10 instance, 9 of 20 clients have a single target per class, yet pooled, the largest target of a class receives on average only 33\% of its flipped labels (8\% on CIFAR-100).}
  \label{fig:spectrum}
\end{figure}

\paragraph{Federated learning.}
Clients $k=1,\dots,M$ hold data $D_k=\{(x_i,\ty_i)\}$ of $n_k$ images with observed labels in $\{1,\dots,C\}$, whose true labels $y_i$ are never observed.
We write $(X,Y,\tY,K)$ for the image, true label, observed label, and client of a random training sample.
The clients train a shared model $\theta$ with FedAvg~\cite{fedavg}. In round $r$ the server sends $\theta^r$ to all clients, each client $k$ trains it for a few local epochs on $D_k$ to obtain $\theta^r_k$, and the server sets
\begin{equation}
  \theta^{r+1}=\sum_{k=1}^{M}\frac{n_k}{n}\,\theta^r_k,\qquad n=\sum_{k=1}^{M}n_k.
  \label{eq:fedavg}
\end{equation}
Images and labels never leave a client, and only models and, in \method, a few aggregates are exchanged (\cref{sec:comm}).

\paragraph{Federated label noise.}
We write $\pi_k(c)=\Prob(Y=c\mid K=k)$ for the class prior of client $k$ and $T_k(j\mid c)=\Prob(\tY=j\mid Y=c,K=k)$ for its class-conditional noise transition matrix~\cite{natarajan2013learning,patrini2017making}.
Federated learning with noisy labels seeks a global model that predicts the true label $Y$ of a test image, learned by FL from the observed labels alone, with no clean labels at any client or at the server.
Existing benchmarks let clients differ in how many labels are wrong, or flip each client's labels only within its local classes~\cite{fedcorr,fednoisy}.
We study the general case of \emph{client-specific label noise}, in which $T_k$ itself varies across clients, in where labels flip as well as in how often.
Clients also differ in $\pi_k$, and we call the pair $(\pi_k,T_k)$ the client's \emph{labeling signature}.
Our synthetic benchmark, the \emph{noise spectrum}, makes the transition matrices differ in where the flips go. Client $k$ draws a sparsity $s_k\sim\mathrm{Beta}(0.5,0.5)$ and sends the flipped labels of each class to $\ell_k=\max\{1,\mathrm{round}((1-s_k)(C-1))\}$ targets, drawn independently for every client and class, and a flipped label picks one of them uniformly.
$\ell_k=1$ is a pair flip and $\ell_k=C-1$ is symmetric noise.
The Beta$(0.5,0.5)$ draw puts most clients near the two ends and the rest in between (\cref{fig:spectrum}).
Two clients with the same classes can thus have different transition matrices, which no single shared matrix describes.
A global transition matrix sees only the pooled flips, which are diffuse even when most clients are concentrated.

\section{Method}
\label{sec:method}

\begin{figure*}[t]
  \centering
  \includegraphics[width=0.95\textwidth]{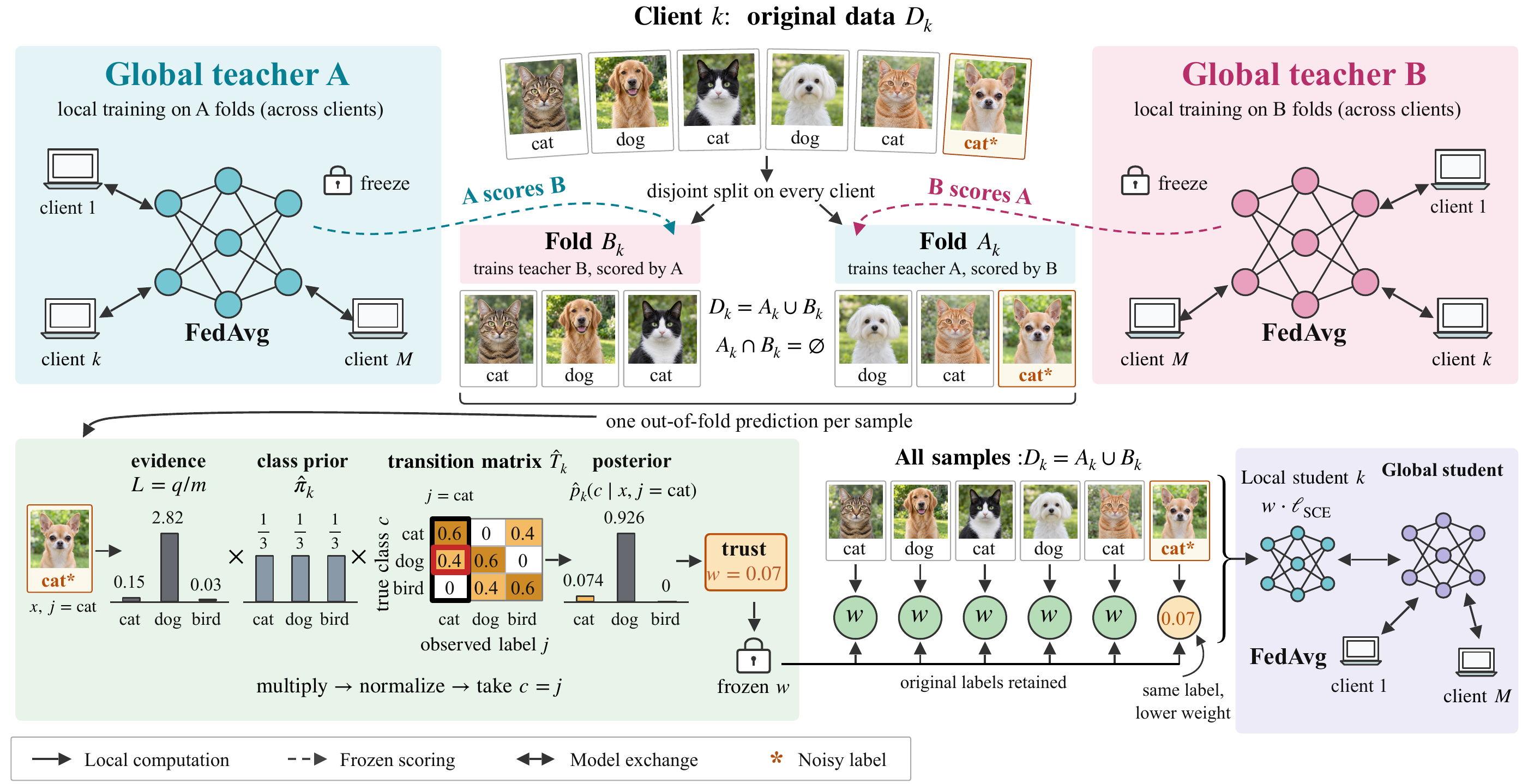}
  \caption{\textbf{\method.}
  \emph{Cross-fitted teachers:} each client $k$ splits $D_k$ into folds $A_k$ and $B_k$; teacher~A is trained with FedAvg on the A~folds of all $M$ clients and scores the B~folds, and vice versa, so no label is scored by a teacher that trained on it.
  \emph{Trust inference:} for image $x$ with observed label $j$, the client turns the calibrated prediction $q$ into evidence $L=q/m$ ($m$: reference prior), multiplies it by its labeling signature, the estimated class prior $\hat\pi_k$ and transition matrix $\hat T_k$, and takes the posterior at $c=j$ as the trust $w$.
  Shown: the wrong label cat* at a client that often writes dogs as cats (black frame: column $j$; red: dog$\to$cat), as a three-class toy computed from \cref{eq:posterior} with uniform priors.
  \emph{Trust-weighted student:} the student keeps every original label and trains with $w\cdot\ell_{\mathrm{SCE}}$ under FedAvg. Photos are illustrative.}
  \label{fig:method}
\end{figure*}

\method estimates and uses label trust in three steps (\cref{fig:method}), after \cref{sec:target} defines the target.
First, two cross-fitted global teachers are trained with FedAvg, each on one fold of every client (\cref{sec:who}).
A label's \emph{in-sample score} is the calibrated probability from the teacher trained on its fold, and its \emph{out-of-fold score} comes from the other teacher.
Second, each client infers a frozen trust from the out-of-fold prediction through its labeling signature (\cref{sec:read}).
Finally, a federated student trains with FedAvg on all original labels weighted by this trust (\cref{sec:use}).

\subsection{Label trust}
\label{sec:target}

For a sample of client $k$ with image $x$ and observed label $j$, define its \emph{trust}
\begin{equation}
  \wstar(x,j,k)=\Prob\big(Y=j\mid X=x,\ \tY=j,\ K=k\big).
  \label{eq:trust}
\end{equation}
It answers only whether the label is right, not what it should be otherwise.
That turns out to be enough for training.

\begin{proposition}[Trust identity]
\label{prop:identity}
For any predictor $f$ and any loss $\ell\ge 0$, with $\Prob(\tY=Y)>0$,
\begin{equation}
  \E\big[\wstar\,\ell(f(X),\tY)\big]=\E\big[\ind\{\tY=Y\}\,\ell(f(X),Y)\big].
\end{equation}
Hence the trust-weighted risk equals the risk on correctly labeled samples, $R_{\wstar}=\Rcor$, where $R_{\wstar}(f)=\E[\wstar\ell(f(X),\tY)]/\E[\wstar]$ and $\Rcor(f)=\E[\ell(f(X),Y)\mid\tY=Y]$.
\end{proposition}
The proof takes two lines (Appendix~\ref{supp:proofs}).
On the event $\{\tY=Y\}$ the two losses coincide, and conditioning on $(X,\tY,K)$ turns the indicator into $\wstar$.
No noise model is involved.
$\Rcor$ equals the clean risk when every label is kept with the same probability, as in our synthetic benchmark, and Appendix~\ref{supp:proofs} bounds the gap otherwise.

\subsection{Cross-fitted teachers}
\label{sec:who}

Each client splits its data into two folds, keeping duplicate images and all slices of each patient in the same fold.
Teacher~A is trained with FedAvg on the A~folds of all clients, teacher~B on the B~folds, for 60 rounds each.
A sample's \emph{out-of-fold prediction} $q(\cdot\mid x)$, its second opinion, comes from the teacher of the other fold.
Teachers and student start from the same encoder, which contrastive learning~\cite{simclr} trains on the clients' images without labels and which helps learning with noisy labels~\cite{c2d}.
The encoder has thus seen every image, but no teacher has been trained on the label of a sample it scores.

Cross-client aggregation provides a second opinion beyond a single client's labeling behavior.
A teacher trained only on one client's complementary fold sees fewer examples and remains exposed to that client's systematic confusions.
Across clients whose errors point to different classes, federated training learns shared visual representations while diluting individual flipping rules (\cref{fig:spectrum}).
Before trust inference, we calibrate each teacher's predictions on held-out noisy labels using a scalar temperature $\tau$ and a fixed reference transition matrix.
Dividing the calibrated prediction $q(c\mid x)$ by its mean $m(c)$ over calibration samples gives the evidence $L(c,x)=q(c\mid x)/m(c)$~\cite{saerens2002adjusting}, which estimates the class likelihood ratio used in \cref{prop:posterior}.
Here $m$ serves as the reference prior, and calibration details are in Appendix~\ref{supp:protocol}.

\subsection{Labeling signatures and trust inference}
\label{sec:read}

\begin{proposition}[Posterior form]
\label{prop:posterior}
Assume (A1) $\tY\perp X\mid(Y,K)$, \ie within a client the noise depends only on the class~\cite{natarajan2013learning,patrini2017making}, and (A2) $X\perp K\mid Y$, \ie clients share the appearance of each class.
For any full-support prior $m$, let $q_m(c\mid x)\propto p(x\mid c)\,m(c)$. Then, whenever the denominator is positive,
\begin{equation}
  \Prob(Y=c\mid x,j,k)=\frac{\frac{q_m(c\mid x)}{m(c)}\,\pi_k(c)\,T_k(j\mid c)}{\sum_{c'}\frac{q_m(c'\mid x)}{m(c')}\,\pi_k(c')\,T_k(j\mid c')},
  \label{eq:posterior}
\end{equation}
and the trust $\wstar$ is its value at $c=j$.
\end{proposition}

\method plugs the out-of-fold evidence and the client's estimated signature $(\hat\pi_k,\hat T_k)$ into \cref{eq:posterior}, which gives $\hat p_k(c\mid x,j)\propto L(c,x)\,\hat\pi_k(c)\,\hat T_k(j\mid c)$.
\method's \emph{trust} is $w=\hat p_k(j\mid x,j)$, an estimate of $\wstar$.
Each client fits its signature from out-of-fold evidence and observed labels by maximum-a-posteriori EM with Dirichlet smoothing~\cite{dawidskene,saerens2002adjusting}.
Inner cross-validation selects the matrix-prior strength $\beta$, and each sample is scored using signature parameters fitted on the other inner folds (Appendix~\ref{supp:protocol}).

To see how the signature shapes trust and why the score alone fails, write $q_c=q(c\mid x)$ and $\alpha_c=q_c/(1-q_j)$ for the teacher's distribution over the alternatives $c\neq j$.

\begin{corollary}[Odds decomposition]
\label{cor:odds}
For $q_j<1$ and a positive denominator, \method's trust satisfies
\begin{equation}
  \frac{w}{1-w}
  =\underbrace{\frac{q_j}{1-q_j}}_{\text{teacher odds}}
  \times
  \underbrace{\frac{\frac{\hat\pi_k(j)}{m(j)}\,\hat T_k(j\mid j)}{\sum_{c\neq j}\alpha_c\,\frac{\hat\pi_k(c)}{m(c)}\,\hat T_k(j\mid c)}}_{\text{client factor }\lambda_k(x,j)}.
  \label{eq:odds}
\end{equation}
\end{corollary}
The identity holds for any input prediction and any fitted signature.
The client factor compares how likely client $k$ is to keep a true $j$ with how likely it is to turn the teacher's plausible alternatives into $j$.
Trust inference raises the teacher's score where $\lambda_k>1$ and lowers it where $\lambda_k<1$.
With the true signature, $\pi_k=m$, and the noise spectrum with flip rate $\rho$, it becomes $\lambda_k=(1-\rho)\,\ell_k/(\rho\,\eta_k)$, where $\eta_k=\sum_{c\neq j}\alpha_c\,\ind\{\text{client }k\text{ flips }c\text{ into }j\}\in[0,1]$ (Appendix~\ref{supp:proofs}).
Three consequences follow.
\begin{enumerate}[leftmargin=*,itemsep=1pt,topsep=2pt,label=(\roman*)]
  \item \textbf{The score alone under-trusts.} Using $q_j$ directly as the trust sets $\lambda_k=1$. At a dispersed client each alternative flips into $j$ at rate only $\rho/\ell_k$, so $\lambda_k$ is large. With $C=100$, $\rho=0.4$, and $\ell_k=99$, $\lambda_k\approx148$, and a label given only $0.05$ by the teacher still has trust $\approx0.89$.
  \item \textbf{Trust inference is client-specific.} Suppose the teacher gives a label $0.05$. At a concentrated client ($\ell_k=1$), $\lambda_k=1.5/\eta_k$. If the teacher's alternatives are classes that this client flips into $j$, then $\lambda_k\approx1.5$ cannot lift the low odds, the trust stays near $0.07$, and the doubt stands. If they are other classes, $\eta_k\approx0$, $\lambda_k$ is huge, and the label is trusted. The same prediction thus leads to opposite conclusions at two clients.
  \item \textbf{Judging does not require naming.} Trust is the posterior mass on the observed label alone. Naming the true class would also require ranking $\alpha_c\,\pi_k(c)\,T_k(j\mid c)/m(c)$ over $c\neq j$, on which a symmetric transition matrix gives no hint beyond the teacher. \method estimates only the trust that training needs.
\end{enumerate}

\paragraph{Trust inference does not discount memorized predictions.}
\Cref{prop:posterior} treats the prediction as evidence about the image alone, and a prediction from a teacher that has trained on $(x_i,\ty_i)$ is not.
The client factor multiplies the teacher odds whatever their origin.
As $q(\ty_i\mid x_i)\to1$ the label gets $w\to1$ for any fixed $\hat\pi_k$, $\hat T_k$, and $m$ with $\hat\pi_k(\ty_i)\,\hat T_k(\ty_i\mid\ty_i)>0$, and wherever $\lambda_k>1$ the inference magnifies the memorized odds of wrong labels.
Cross-fitting removes this path, which is why the prediction must come from a teacher that never trained on the label.

\begin{table*}[t]
  \centering
  \footnotesize
  \setlength{\tabcolsep}{5.1pt}
  \caption{\textbf{Main results.} Last-10-round means over three training seeds, in \% (standard deviations in Appendix~\ref{supp:results}). Acc: accuracy; Rec: macro recall (class-imbalanced ICH); Prec and F1: macro. \textbf{Bold}: best; \underline{underline}: second best. Baselines use their original initialization.}
  \label{tab:main}
  \begin{tabular}{@{}lccccccccccccccc@{}}
    \toprule
    & \multicolumn{9}{c}{Synthetic client-specific noise (noise spectrum, 40\% flips)} & \multicolumn{6}{c}{Real-world label noise} \\
    \cmidrule(lr){2-10}\cmidrule(l){11-16}
    & \multicolumn{3}{c}{CIFAR-10} & \multicolumn{3}{c}{CIFAR-100} & \multicolumn{3}{c}{ICH} & \multicolumn{3}{c}{CIFAR-10N} & \multicolumn{3}{c}{CIFAR-100N} \\
    \cmidrule(lr){2-4}\cmidrule(lr){5-7}\cmidrule(lr){8-10}\cmidrule(lr){11-13}\cmidrule(l){14-16}
    Method & Acc & Prec & F1 & Acc & Prec & F1 & Rec & Prec & F1 & Acc & Prec & F1 & Acc & Prec & F1 \\
    \midrule
    FedAvg~\cite{fedavg} & 72.77 & 73.27 & 72.64 & 40.67 & 40.44 & 40.18 & 26.68 & 27.39 & 24.51 & 69.11 & 69.82 & 68.92 & 50.91 & 51.79 & 49.94 \\
    FedProx~\cite{fedprox} & 71.84 & 72.06 & 71.69 & 40.23 & 39.70 & 39.71 & 27.08 & 27.61 & 25.71 & 67.32 & 67.60 & 67.10 & 50.08 & 50.35 & 49.10 \\
    SCE~\cite{sce} & 81.80 & 82.28 & 81.63 & 54.61 & 54.74 & 54.26 & 29.52 & 29.61 & 28.45 & \underline{79.50} & \underline{79.95} & \underline{79.34} & 53.95 & 54.49 & 52.80 \\
    Co-teaching~\cite{coteaching} & 75.58 & 76.10 & 75.39 & 48.62 & 48.29 & 48.02 & 27.30 & 27.96 & 25.84 & 75.31 & 75.85 & 75.16 & 54.82 & 53.38 & 52.97 \\
    INCV~\cite{incv} & 72.86 & 76.28 & 72.50 & 42.35 & 41.18 & 39.91 & 31.87 & 40.12 & 31.61 & 75.91 & 77.11 & 75.49 & 44.30 & 43.59 & 41.49 \\
    CL~\cite{confidentlearning} & 78.88 & 79.78 & 78.87 & 22.21 & 32.80 & 21.76 & 28.87 & 29.38 & 28.08 & 77.15 & 77.87 & 76.96 & 34.21 & 42.65 & 32.78 \\
    \midrule
    FedCorr~\cite{fedcorr} & 75.95 & 79.72 & 74.98 & 51.79 & 52.30 & 50.73 & 23.43 & 25.65 & 18.92 & 72.76 & 74.86 & 72.41 & 54.67 & 57.69 & 53.35 \\
    FedNoRo~\cite{fednoro} & 66.92 & 68.92 & 66.51 & 42.52 & 49.30 & 43.41 & 29.35 & 34.00 & 28.99 & 65.53 & 65.95 & 65.24 & 51.54 & 51.81 & 50.54 \\
    FedRN~\cite{fedrn} & 77.71 & 78.28 & 77.51 & 46.35 & 46.25 & 45.87 & 28.31 & 28.81 & 27.30 & 74.37 & 75.33 & 74.24 & 53.46 & 54.68 & 52.52 \\
    FedELC~\cite{fedelc} & 76.54 & 76.76 & 76.35 & 39.52 & 39.53 & 38.51 & 29.29 & 29.64 & 27.77 & 70.16 & 70.48 & 69.97 & 48.81 & 48.40 & 47.76 \\
    FedBeat~\cite{fedbeat} & 76.69 & 79.50 & 77.25 & 48.64 & 46.03 & 45.96 & 21.50 & 9.55 & 6.91 & 66.60 & 68.87 & 67.15 & 52.55 & 52.72 & 50.78 \\
    FedNed~\cite{fedned} & 72.76 & 73.27 & 72.61 & 40.70 & 40.51 & 40.22 & 27.34 & 27.99 & 25.42 & 69.10 & 69.78 & 68.96 & 50.83 & 51.63 & 49.80 \\
    FedEFC~\cite{fedefc} & 66.07 & 66.28 & 65.94 & 35.75 & 35.26 & 35.29 & 29.65 & 29.30 & 28.16 & 61.97 & 62.44 & 61.86 & 46.82 & 46.57 & 45.90 \\
    FedLTF~\cite{fedltf} & 47.53 & 57.74 & 45.18 & 21.79 & 44.31 & 18.88 & 30.46 & 37.29 & 30.56 & 60.87 & 66.40 & 60.47 & 23.43 & 46.87 & 21.05 \\
    FedRG~\cite{fedrg} & \underline{83.70} & \underline{84.14} & \underline{83.61} & \underline{64.55} & \underline{65.05} & \underline{64.17} & \underline{43.14} & \underline{42.50} & \underline{42.00} & 78.06 & 78.52 & 77.97 & \underline{58.55} & \underline{58.61} & \underline{57.40} \\
    \midrule
    \oursrow \method (ours) & \textbf{87.84} & \textbf{87.91} & \textbf{87.74} & \textbf{68.03} & \textbf{68.45} & \textbf{67.67} & \textbf{49.31} & \textbf{48.66} & \textbf{48.54} & \textbf{83.54} & \textbf{83.97} & \textbf{83.36} & \textbf{61.39} & \textbf{62.54} & \textbf{60.04} \\
    \textit{vs.\ Best} & \textit{+4.14} & \textit{+3.77} & \textit{+4.13} & \textit{+3.48} & \textit{+3.40} & \textit{+3.50} & \textit{+6.17} & \textit{+6.16} & \textit{+6.54} & \textit{+4.04} & \textit{+4.02} & \textit{+4.02} & \textit{+2.84} & \textit{+3.93} & \textit{+2.64} \\
    \bottomrule
  \end{tabular}
\end{table*}

\subsection{Trust-weighted training}
\label{sec:use}

The student trains with FedAvg on every original label.
Client $k$ minimizes
\begin{equation}
  \frac{1}{|\mathcal B|}\sum_{i\in\mathcal B}\frac{w_i}{\bar w}\,\ell_{\mathrm{SCE}}\big(p_\theta(x_i),\ty_i\big)
  \label{eq:student}
\end{equation}
on each minibatch $\mathcal B$, where $\bar w=\sum_k\sum_{i\in D_k}w_i/n$ is the mean trust over all samples and $\ell_{\mathrm{SCE}}$ is the symmetric cross-entropy with weights $0.1$ (CE) and $1$ (reverse CE)~\cite{sce}.
The trust is computed once and frozen.
No label is changed, no sample is removed, and the server aggregates by \cref{eq:fedavg}.
With the federation-wide $\bar w$, the $n_k/n$-weighted sum of the client losses is the trust-weighted risk of the whole federation, the sample form of the trust-weighted risk in \cref{prop:identity}, with $w$ in place of $\wstar$.
Global normalization preserves differences in clients' mean inferred trust in this objective, without a separate client-selection step.
Trust and the robust loss act on different things, since trust is a fixed coefficient from outside the student, whereas the SCE gradient scales with the student's own prediction (Appendix~\ref{supp:protocol}).
Appendix~\ref{supp:proofs} relates errors in the estimated trust to deviations from the risk on correctly labeled samples.

\subsection{Communication and cost}
\label{sec:comm}

Besides model parameters, clients send three small aggregates. They are the values of the temperature objective at the temperatures queried by a scalar optimizer, a $C$-dimensional mean prediction for the reference prior, and the sum of their trust values for $\bar w$.
Labels, signatures, and per-sample trust stay on the client.
The teachers add about 54 full-data rounds of computation (two 60-round runs on about 45\% of the data) and 120 communication rounds.

\section{Experiments}
\label{sec:exp}

\subsection{Setup}
\label{sec:setup}

\paragraph{Benchmarks and protocol.}
Synthetic client-specific noise follows the noise spectrum of \cref{sec:setting}, with 40\% of each client's labels flipped, on CIFAR-10/100~\cite{cifar} and on the five-class intracranial hemorrhage CT benchmark (ICH) of FedNoRo~\cite{fednoro}, built from the RSNA challenge data~\cite{ich}, which we split so that no patient appears on both sides.
Real-world noise comes from the human annotations of CIFAR-10N/100N~\cite{cifarn}, whose errors follow human confusions that recur across clients.
Twenty clients receive a Dirichlet(0.5) class split~\cite{hsu2019measuring}, all take part in every round, and models are ResNet-18~\cite{resnet}.
We report the mean over the last 10 rounds, averaged over three training seeds for each data split and noise draw.
All training details are in Appendix~\ref{supp:benchmarks}.
Code will be made publicly available.

\paragraph{Baselines.}
We compare with fifteen baselines in two groups, general FL and noise-robust learning run in FL (FedAvg~\cite{fedavg}, FedProx~\cite{fedprox}, SCE~\cite{sce}, Co-teaching~\cite{coteaching}, INCV~\cite{incv}, Confident Learning (CL)~\cite{confidentlearning}), and federated noisy-label learning (FedCorr~\cite{fedcorr}, FedNoRo~\cite{fednoro}, FedRN~\cite{fedrn}, FedELC~\cite{fedelc}, FedBeat~\cite{fedbeat}, FedNed~\cite{fedned}, FedEFC~\cite{fedefc}, FedLTF~\cite{fedltf}, FedRG~\cite{fedrg}).
We federate INCV and CL with FedAvg, so every model they train is trained across all clients and every label is screened on the client that holds it.
Each baseline follows the initialization of its original method and gets the same student budget as \method (Appendix~\ref{supp:baselines}).

\begin{figure}[t]
  \centering
  \includegraphics[width=\linewidth]{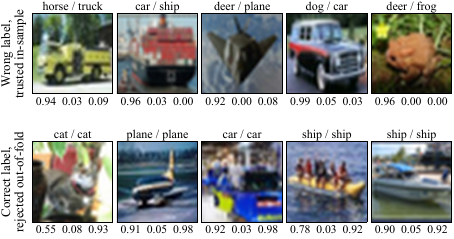}
  \caption{\textbf{The two failures, on real images} (CIFAR-10). Above each image: observed label / true class. Below: in-sample and out-of-fold scores used alone, and \method trust. \emph{Top:} wrong labels trusted ($\ge0.9$) by the teacher trained on them, while the out-of-fold score and \method reject them ($<0.1$). \emph{Bottom:} correct labels that the out-of-fold score alone rejects ($<0.1$) and \method trusts ($\ge0.9$). Random draws among the qualifying labels; more in Appendix~\ref{supp:mechanism}.}
  \label{fig:examples}
\end{figure}

\subsection{Main results}
\label{sec:main}

\method is best on all five benchmarks and all three metrics (\cref{tab:main}).
Under synthetic client-specific noise it exceeds the strongest baseline by 4.14 points on CIFAR-10, 3.48 on CIFAR-100, and 6.17 points of macro recall on ICH, and under real-world noise by 4.04 on CIFAR-10N and 2.84 on CIFAR-100N.
\method's standard deviation on the primary metric is at most 0.51 points, and its worst seed beats the best seed of every baseline on all five benchmarks.
\method's final accuracy also exceeds the best 10-round window of every baseline on all four CIFAR benchmarks (Appendix~\ref{supp:results}).
With identical self-supervised initialization, \method still exceeds the strongest compared baseline by 1.60 and 3.48 points on CIFAR-10/100, respectively (Appendix~\ref{supp:baselines}).

\subsection{Ablation study}
\label{sec:ablation}

\begin{table}[t]
  \centering
  \footnotesize
  \setlength{\tabcolsep}{3.2pt}
  \newcommand{\dd}[1]{{\scriptsize\textcolor{gray}{#1}}}
  \caption{\textbf{Prediction source $\times$ trust inference} (top) and other replacements (bottom); accuracy, \%. Gray values show changes relative to \method. \emph{In-sample:} scored by the teacher that trained on the label. \emph{Score only:} the calibrated score is the trust. \emph{Signature:} trust is inferred through the client's labeling signature (\cref{eq:posterior}). \emph{Shared signature:} one signature for all clients, the reference prior $m$ and a jointly fitted transition matrix, replaces the client's own. \emph{No trust:} weight 1. \emph{Relabel:} labels with trust below 0.5 take the posterior's most likely class. Standard deviations are at most 0.37 (Appendix~\ref{supp:variants}).}
  \begin{tabular}{@{}lcccc@{}}
    \toprule
    & \multicolumn{2}{c}{CIFAR-10} & \multicolumn{2}{c}{CIFAR-100} \\
    \cmidrule(lr){2-3}\cmidrule(l){4-5}
    Prediction & Score only & Signature & Score only & Signature \\
    \midrule
    In-sample & 86.23 & 85.49 & 64.96 & 63.30 \\
    Out-of-fold & 85.78 & \cellcolor[gray]{0.93}\textbf{87.84} & 61.41 & \cellcolor[gray]{0.93}\textbf{68.03} \\
    \midrule
    No trust ($w\equiv1$) & \multicolumn{2}{c}{82.14 \dd{$-$5.70}} & \multicolumn{2}{c}{62.41 \dd{$-$5.62}} \\
    Shared signature & \multicolumn{2}{c}{86.62 \dd{$-$1.22}} & \multicolumn{2}{c}{66.97 \dd{$-$1.06}} \\
    Relabel & \multicolumn{2}{c}{86.32 \dd{$-$1.52}} & \multicolumn{2}{c}{64.22 \dd{$-$3.81}} \\
    CE instead of SCE & \multicolumn{2}{c}{84.92 \dd{$-$2.92}} & \multicolumn{2}{c}{63.81 \dd{$-$4.22}} \\
    \bottomrule
  \end{tabular}
  \label{tab:ablation}
\end{table}

\begin{table}[t]
  \centering
  \footnotesize
  \setlength{\tabcolsep}{1.2pt}
  \caption{\textbf{Scoring the same labels five ways.} Labels scored by both teachers, threshold 0.5; terms as in \cref{tab:ablation}. Acc$_\text{w}$: share of wrong labels accepted; Rej$_\text{c}$: share of correct labels rejected (\%); AUROC: separation of correct from wrong labels. Calibration is in \cref{fig:reliability} and weight shares are in Appendix~\ref{supp:mechanism}.}
  \begin{tabular}{@{}lcccccc@{}}
    \toprule
    & \multicolumn{3}{c}{CIFAR-10} & \multicolumn{3}{c}{CIFAR-100} \\
    \cmidrule(lr){2-4}\cmidrule(l){5-7}
    Score & Acc$_\text{w}\!\downarrow$ & Rej$_\text{c}\!\downarrow$ & AUROC$\uparrow$ & Acc$_\text{w}\!\downarrow$ & Rej$_\text{c}\!\downarrow$ & AUROC$\uparrow$ \\
    \midrule
    In-sample, score only & 26.2 & 14.5 & .883 & 25.2 & 19.4 & .864 \\
    In-sample, signature & 62.7 & 0.2 & .951 & 96.7 & 0.0 & .920 \\
    Out-of-fold, score only & 2.8 & 44.3 & .917 & 0.1 & 70.9 & .934 \\
    Out-of-fold, shared & 14.8 & 15.9 & .924 & 12.3 & 17.9 & .924 \\
    \oursrow \method & 7.3 & 8.8 & \textbf{.977} & 8.4 & 11.7 & \textbf{.965} \\
    \bottomrule
  \end{tabular}
  \label{tab:scores}
\end{table}

\Cref{tab:ablation} keeps the teachers and the student training fixed and replaces one design at a time.

\emph{Prediction source and trust inference.}
The top of \cref{tab:ablation} crosses where the prediction comes from with how trust is obtained.
From in-sample scores used alone (86.23\% and 64.96\%), changing either choice by itself hurts.
Out-of-fold scores used alone drop accuracy to 85.78\% ($-0.45$) and 61.41\% ($-3.55$) because they under-trust hard clean labels.
Inferring trust from in-sample scores drops it to 85.49\% ($-0.74$) and 63.30\% ($-1.66$) because the inference reinforces memorized errors (\cref{sec:read}).
Changing both reaches 87.84\% and 68.03\%, which is $+1.61$ and $+3.07$ over in-sample scores used alone and $+2.06$ and $+6.62$ over out-of-fold scores used alone.
The gain is largest on CIFAR-100, where dispersed flips make $\lambda_k$ largest (\cref{cor:odds}).

\emph{How it is used.}
Using the trust to relabel costs 1.52 and 3.81 points.
Relabeling must also name a replacement, which the trust never needs (consequence (iii) of \cref{cor:odds}).
A hard mask at 0.5, which keeps the original labels, comes closer ($-0.52$ and $-0.99$).
Training the student with CE instead of SCE costs 2.92 and 4.22 points.
Trust adds 13.28 and 13.17 points to a CE student and 5.70 and 5.62 to an SCE student (Appendix~\ref{supp:variants}), so each helps on top of the other.

\subsection{Analysis}
\label{sec:analysis}

\begin{figure}[t]
  \centering
  \includegraphics[width=\linewidth]{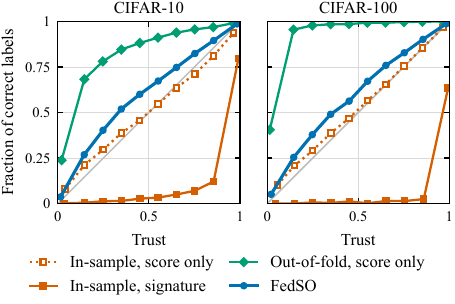}
  \caption{\textbf{From prediction scores to label trust.} Scores are grouped into ten equal-width bins. Each point shows the mean score and fraction of correct labels, averaged over three seeds. The diagonal indicates perfect calibration. Out-of-fold scores used alone lie above it (under-trust), while trust inferred from in-sample scores lies below it (over-trust). \method stays close to the diagonal on both datasets.}
  \label{fig:reliability}
\end{figure}

\begin{figure}[t]
  \centering
  \includegraphics[width=\linewidth]{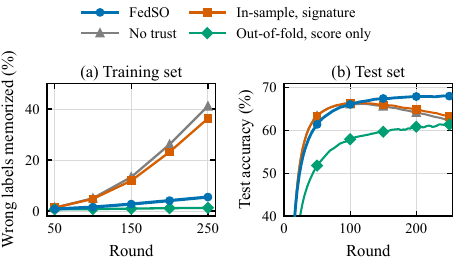}
  \caption{\textbf{What the student ends up believing} (CIFAR-100).
  (a)~Share of wrong training labels that the student memorizes, measured every 50 rounds.
  (b)~Test accuracy (10-round mean).
  With trust inferred from in-sample scores, the student memorizes almost as many wrong labels as without trust, and its test accuracy declines after round 100. With the out-of-fold score alone, memorization stays near zero but accuracy stays low. \method avoids both.}
  \label{fig:beliefs}
\end{figure}

\paragraph{Trust quality.}
\Cref{tab:scores} scores the same labels in five ways.
\method ranks labels best, with AUROC .977 and .965, and is calibrated, with expected calibration error (ECE, lower is better)~\cite{guo2017calibration} of .031 and .040 (\cref{fig:reliability}).
The out-of-fold score alone rejects 44.3\% and 70.9\% of the correct labels and is badly calibrated (ECE .248 and .407).
It even gives wrong labels less weight than \method (1.5\% \vs 9.1\% on CIFAR-100), yet its student is 6.62 points worse, because it starves the hard correct labels.
Trust inference on in-sample scores fails the other way.
It accepts 62.7\% and 96.7\% of the wrong labels, up from 26.2\% and 25.2\% for the in-sample score alone, as \cref{sec:read} predicts.
The same inference thus gives the best trust from out-of-fold predictions and the worst from in-sample ones.
\Cref{fig:examples} shows both failures on real images.
A fire truck labeled ``horse'' gets 0.94 from the in-sample teacher, a cat labeled ``cat'' gets 0.08 from the out-of-fold score, and \method gets both right.
Whereas uniform weights give wrong labels 40\% of the supervision, \method gives them 7.1\% and 9.1\% on CIFAR-10/100.

\paragraph{Training dynamics.}
\Cref{fig:beliefs} follows the student through training on CIFAR-100.
A student that trusts every label memorizes 41.1\% of the wrong labels by round 250.
Trust inferred from in-sample scores lowers this only to 36.3\%, and test accuracy falls from 66.38 at round 100 to 63.30.
With the out-of-fold score alone the student memorizes almost no wrong labels (1.2\%) but learns only 73.3\% of the correctly labeled samples.
The \method student memorizes 5.5\%, learns 94.7\%, and its test accuracy rises from 66.09 to 68.03.
The two failures of \cref{fig:teaser} thus reappear in training and explain the accuracy gaps of \cref{tab:ablation}.
CIFAR-10 shows the same two failures with smaller gaps (Appendix~\ref{supp:mechanism}).

\begin{figure}[t]
  \centering
  \includegraphics[width=\linewidth]{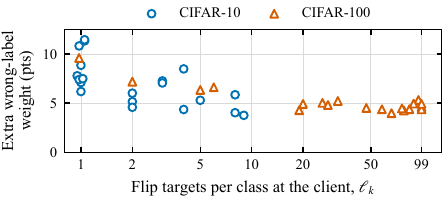}
  \caption{\textbf{Where the client's own signature matters.} Each marker is one client (20 per dataset). Horizontal: number of flip targets per class at that client, $\ell_k$ (log scale). Vertical: how many more points of the total weight fall on wrong labels when the client's own signature is replaced by a shared one.}
  \label{fig:locality}
\end{figure}

\paragraph{Locality.}
Replacing the client's signature by a shared one costs 1.22 and 1.06 points (\cref{tab:ablation}), accepts more wrong labels, rejects more correct ones, and ranks labels worse (AUROC .924 on both datasets, \cref{tab:scores}).
Client by client (\cref{fig:locality}), the client's own signature lowers the share of weight on wrong labels for all 20 clients of both datasets, by 3.8 to 11.4 points on CIFAR-10 and 4.0 to 9.6 on CIFAR-100.
The reduction is largest where noise is concentrated, 8.71 \vs 5.62 points on CIFAR-10 and 9.59 \vs 4.97 on CIFAR-100 for clients with $\ell_k=1$ against the rest.
Separating the signature's two components shows that they act on different errors (Appendix~\ref{supp:variants}).
Sharing the matrix mainly increases wrong-label weight at clients with concentrated flips ($\ell_k=1$), matching consequence (ii) of \cref{cor:odds}.
Sharing the prior instead increases rejection of correct labels in locally frequent classes and acceptance of wrong labels in locally rare classes (Appendix~\ref{supp:variants}).

\paragraph{Shared flip targets.}
A controlled test keeps the client partition, corrupted samples, and each $\ell_k$ fixed, but draws targets from one shared ranking per class (Appendix~\ref{supp:shared}).
On CIFAR-10, the out-of-fold teacher's agreement with wrong labels rises from 4.7\% to 11.2\%, yet \method loses only 0.54 accuracy points, against 2.05 for the unweighted SCE student.
Trust retains AUROC .960 and .961 on CIFAR-10/100, and \method's CIFAR-100 accuracy changes by less than 0.01 points.

\paragraph{Real-world noise.}
Human annotations likewise contain confusions that recur across clients.
On CIFAR-10N and CIFAR-100N the in-sample teacher memorizes 64.2\% and 68.1\% of the wrong labels it trained on, while the out-of-fold teacher predicts 13.0\% and 12.4\% of them.
This rate is higher than under synthetic noise (4.9\% and 0.6\%).
Plausible human confusions shared across clients are harder to dilute through aggregation.
Accordingly, \method's trust is less sharp on real-world noise (AUROC .944 and .889), yet it still leads by 4.04 and 2.84 points (\cref{tab:main}).

\subsection{Number of clients and cost}
\label{sec:cost}

\begin{figure}[t]
  \centering
  \includegraphics[width=\linewidth]{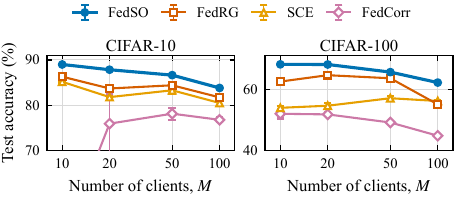}
  \caption{\textbf{Number of clients.} Test accuracy (mean and standard deviation over three seeds) with the CIFAR training set split among $M$ clients, each $M$ with its own class split and noise draw.}
  \label{fig:clients}
\end{figure}

\paragraph{Number of clients.}
\Cref{fig:clients} splits the CIFAR training images among 10, 50, or 100 clients instead of 20.
\method leads FedRG, SCE, and FedCorr at every size, by 2.01 to 6.00 points, and its worst seed beats the best seed of every baseline.
At $M=100$, where the smallest CIFAR-10 client holds 116 images, the trust keeps AUROC .959 and .953 on CIFAR-10/100 and puts 10.1\% and 10.7\% of the total weight on wrong labels (Appendix~\ref{supp:clients}).

\paragraph{Teachers.}
Students trained with teacher checkpoints at 20, 40, and 60 rounds differ by at most 0.12 points, so 20 rounds suffice. This cuts the teacher overhead to about 18 full-data rounds and 40 communication rounds (Appendix~\ref{supp:variants} and~\ref{supp:cost}).

\paragraph{Client side.}
The local EM works on a $C\times C$ transition matrix and a $C$-dimensional prior per client and adds no communication beyond the aggregates of \cref{sec:comm}.
The temperature $\tau$, strength $\beta$, and client signatures are all fitted on held-out noisy labels, never on test labels.

\section{Conclusion}
\label{sec:conclusion}

Whether a label deserves trust depends on both the prediction of a model never trained on it and the labeling behavior of the client that wrote it.
\method turns this principle into a federated procedure.
Cross-fitted global teachers give every sample an out-of-fold second opinion, each client infers trust from it through its private labeling signature, and a student trains on all original labels weighted by this trust.
Clients exchange only models and three small aggregates.
Because weighting judges a label without naming its replacement, \method stays effective across the noise spectrum, from concentrated confusions to dispersed flips.
It outperforms fifteen baselines on five benchmarks, keeps its lead with 10 to 100 clients on CIFAR, and on CIFAR-100 cuts student memorization from 36.3\% with trust inferred from in-sample scores to 5.5\%.
Beyond FL, the same recipe applies wherever labels come from sources that cannot be cross-checked, and the second opinion can come from any model not trained on the labels.

{
    \small
    \bibliographystyle{ieeenat_fullname}
    \bibliography{main}
}

\clearpage
\twocolumn[{\centering\Large\textbf{\thetitle}\\\vspace{0.5em}Appendix\\\vspace{1.0em}}]
\appendix
\crefalias{section}{appendix}\crefalias{subsection}{appendix}
\crefname{appendix}{Appendix}{Appendices}\Crefname{appendix}{Appendix}{Appendices}
\setcounter{topnumber}{3}\setcounter{dbltopnumber}{2}\setcounter{totalnumber}{5}
\renewcommand{\topfraction}{0.9}\renewcommand{\dbltopfraction}{0.9}\renewcommand{\bottomfraction}{0.6}
\renewcommand{\textfraction}{0.08}\renewcommand{\floatpagefraction}{0.8}\renewcommand{\dblfloatpagefraction}{0.8}
\makeatletter\setlength{\@fptop}{0pt}\setlength{\@fpsep}{10pt plus 1fil}\setlength{\@fpbot}{0pt plus 1fil}\setlength{\@dblfptop}{0pt}\setlength{\@dblfpsep}{10pt plus 1fil}\setlength{\@dblfpbot}{0pt plus 1fil}\makeatother

\section{Background, setting, and notation}
\label{supp:background}

\paragraph{Federated learning in brief.}
$M$ clients hold private data and train a shared model $\theta$ with FedAvg~\cite{fedavg}.
In round $r$ the server sends $\theta^r$ to all clients, each client $k$ trains it for a few local epochs on its own data $D_k$ of $n_k$ samples to obtain $\theta^r_k$, and the server sets
\begin{equation}
  \theta^{r+1}=\sum_{k=1}^{M}\frac{n_k}{n}\,\theta^r_k,\qquad n=\sum_{k=1}^{M}n_k.
  \label{eq:supp-fedavg}
\end{equation}
Images and labels never leave a client; only models and, in \method, a few aggregates are exchanged (\cref{supp:protocol}).
Client data are not identically distributed; we simulate this with a Dirichlet split of each class over the clients, where a smaller concentration gives more skewed clients~\cite{hsu2019measuring}.

\paragraph{Label noise in brief.}
Label noise is commonly described by a transition matrix $T(j\mid c)$, the probability that an image of class $c$ is labeled $j$.
Symmetric noise spreads flips evenly over all other classes, pair-flip noise sends every class to one fixed class, and instance-dependent noise lets the flip depend on the image~\cite{natarajan2013learning,patrini2017making,fedbeat}.
Federated noisy-label methods fall into four groups (\cref{sec:related}). Client-level methods detect noisy clients and treat them differently~\cite{fedcorr,fednoro,fedned}, sample-level methods select, relabel, or down-weight samples~\cite{fedrn,fedelc,feddiv}, correction methods estimate transition matrices~\cite{fedbeat,fedefc}, and two-stage methods pretrain without labels and then fine-tune robustly~\cite{fedltf,fedrg}.

\paragraph{Setting.}
Client $k$ holds $D_k=\{(x_i,\ty_i)\}$ with observed labels in $\{1,\dots,C\}$; the true labels $y_i$ are never observed.
$(X,Y,\tY,K)$ denote the image, true label, observed label, and client of a random training sample.
The class prior of client $k$ is $\pi_k(c)=\Prob(Y=c\mid K=k)$, and its class-conditional noise transition matrix is $T_k(j\mid c)=\Prob(\tY=j\mid Y=c,K=k)$; together they form the client's \emph{labeling signature}.
Label noise is \emph{client-specific} when $T_k$ varies across clients; our synthetic \emph{noise spectrum} lets clients differ in how many classes each class flips into (\cref{supp:benchmarks}).
The \emph{trust} of a sample with image $x$ and observed label $j$ at client $k$ is the probability that its label is correct, $\wstar(x,j,k)=\Prob(Y=j\mid X=x,\tY=j,K=k)$.

\paragraph{\method in brief.}
Each client splits $D_k$ into two folds $A_k$ and $B_k$.
Teacher~A is trained with FedAvg on the A~folds of all clients and teacher~B on the B~folds, so each sample has an \emph{in-sample} prediction from the teacher that trained on it and an \emph{out-of-fold} prediction $q(\cdot\mid x)$, its \emph{second opinion}, from the other teacher.
Each teacher's temperature is fitted on held-out noisy labels through a fixed reference matrix $T_0$ (diagonal $0.5$, off-diagonal $0.5/(C-1)$), and the reference prior $m$ is its mean calibrated prediction.
Client $k$ turns the second opinion into evidence $L(c,x)=q(c\mid x)/m(c)$, estimates its labeling signature $(\hat\pi_k,\hat T_k)$ from its own data by MAP-EM, and infers the posterior
\begin{equation}
  \hat p_k(c\mid x,j)=\frac{L(c,x)\,\hat\pi_k(c)\,\hat T_k(j\mid c)}{\sum_{c'}L(c',x)\,\hat\pi_k(c')\,\hat T_k(j\mid c')},
  \label{eq:supp-read}
\end{equation}
whose value at the observed label, $w=\hat p_k(j\mid x,j)$, is \method's trust.
The student trains with FedAvg on all original labels with the loss $(w/\bar w)\,\ell_{\mathrm{SCE}}$, where $\bar w$ is the mean trust over all $n$ samples.

\paragraph{Scores compared in the diagnostics.}
The \emph{in-sample score} and the \emph{out-of-fold score} of a label are the calibrated probabilities that the in-sample and out-of-fold predictions assign to the observed label.
A score used alone (\emph{score only}) is the trust itself.
Trust inference through the client's \emph{signature} passes the prediction through \cref{eq:supp-read} with that client's estimates, and \method applies it to out-of-fold predictions.
A \emph{shared signature} is one signature for all clients instead, the reference prior $m$ and a jointly fitted transition matrix.
At a threshold of 0.5, Acc$_\text{w}$ is the share of wrong labels accepted and Rej$_\text{c}$ the share of correct labels rejected; AUROC measures how well the trust separates correct from wrong labels; \emph{wrong share} is the fraction of the total trust carried by wrong labels; \emph{clean mean} is the mean trust of correct labels; and ECE is the expected calibration error over 10 equal-width bins~\cite{guo2017calibration}.

\begin{table}[!htb]
  \centering\footnotesize
  \setlength{\tabcolsep}{3pt}
  \caption{\textbf{Notation.}}
  \label{tab:supp-notation}
  \begin{tabular}{@{}l p{0.70\linewidth}@{}}
    \toprule
    Symbol & Meaning \\
    \midrule
    $M$, $C$ & number of clients, number of classes \\
    $D_k$, $n_k$, $n$ & data of client $k$, its size, $n=\sum_k n_k$ \\
    $\theta^r$, $\theta^r_k$ & global model in round $r$, local model of client $k$ \\
    $(X,Y,\tY,K)$ & image, true label, observed label, client \\
    $x$, $j$, $c$ & an image, its observed label, a candidate class \\
    $\pi_k(c)$ & class prior of client $k$, $\Prob(Y=c\mid K=k)$ \\
    $T_k(j\mid c)$ & transition matrix of client $k$, $\Prob(\tY=j\mid Y=c,K=k)$ \\
    $T_0$ & reference matrix, diagonal $0.5$ \\
    $q(c\mid x)$ & calibrated out-of-fold prediction (second opinion) \\
    $q_m(c\mid x)$ & Bayes posterior under prior $m$, $\propto p(x\mid c)\,m(c)$ \\
    $m(c)$ & reference prior, mean calibrated prediction \\
    $L(c,x)$ & evidence, $q(c\mid x)/m(c)$ \\
    $\hat\pi_k$, $\hat T_k$ & client $k$'s estimated signature (prior and matrix) \\
    $\wstar(x,j,k)$ & trust, $\Prob(Y=j\mid X=x,\tY=j,K=k)$ \\
    $w$, $\hw$ & \method's trust; any estimate of $\wstar$ \\
    $\bar w$ & mean trust over all $n$ samples \\
    $\ell$, $B$ & loss, its upper bound \\
    $R_w(f)$ & weighted risk, $\E[w\,\ell(f(X),\tY)]/\E[w]$ \\
    $\Rcor(f)$ & risk on correct labels, $\E[\ell(f(X),Y)\mid\tY=Y]$ \\
    $R_{\mathrm{clean}}(f)$ & clean risk, $\E[\ell(f(X),Y)]$ \\
    $\nu$, $\bar\nu$ & probability that a label is kept, its mean \\
    $q_c$, $\alpha_c$ & $q(c\mid x)$ ($q_m(c\mid x)$ in the ideal case); teacher's distribution over $c\neq j$, $q_c/(1-q_j)$ \\
    $\lambda_k$ & client factor in the odds decomposition \\
    $\hat U$, $U^\star$ & actual and ideal unnormalized posteriors of a sample \\
    $r_c(x)$ & likelihood ratio $p(x\mid c)/p_g(x)$ \\
    $v_k(j)$ & trust from client statistics alone, $\Prob(Y=j\mid\tY=j,K=k)$ \\
    $\eta_k$ & teacher mass on classes that client $k$ flips into $j$ \\
    $\rho$ & share of each client's labels that are flipped ($0.4$) \\
    $s_k$, $\ell_k$ & sparsity of client $k$, its flip targets per class \\
    $\tau$, $\beta$ & temperature, matrix pseudo-count strength \\
    \bottomrule
  \end{tabular}
\end{table}

\paragraph{Roadmap.}
\Cref{tab:supp-notation} lists the notation.
\Cref{supp:proofs} proves the statements of \cref{sec:method} and bounds how trust estimation errors affect the student's risk.
\Cref{supp:protocol} gives the algorithm and every implementation detail, and \cref{supp:benchmarks} the experimental setup and benchmarks.
\Cref{supp:baselines} describes the baselines, their adaptations, and unified initialization.
\Cref{supp:results} reports the full results, \cref{supp:clients,supp:shared} vary the number of clients and how errors are shared, and \cref{supp:variants} covers design variants.
\Cref{supp:mechanism} collects score calibration, the estimated transition matrices, qualitative examples, locality, and real-world noise.
\Cref{supp:cost,supp:limitations} cover cost and limitations.

\section{Proofs}
\label{supp:proofs}

\paragraph{Proposition 1 (trust identity).}
\emph{For any predictor $f$ and loss $\ell\ge0$, with $\Prob(\tY=Y)>0$,}
\begin{equation}
  \E[\wstar\,\ell(f(X),\tY)]=\E[\ind\{\tY=Y\}\,\ell(f(X),Y)],
\end{equation}
\emph{and $R_{\wstar}(f)=\Rcor(f)$.}

\noindent\emph{Proof.}
On $\{\tY=Y\}$ we have $\ell(f(X),Y)=\ell(f(X),\tY)$, so the right-hand side equals $\E[\ind\{\tY=Y\}\,\ell(f(X),\tY)]=\E\big[\ell(f(X),\tY)\,\E[\ind\{Y=\tY\}\mid X,\tY,K]\big]$, and the inner expectation is $\wstar(X,\tY,K)$.
Taking $\ell\equiv1$ gives $\E[\wstar]=\Prob(\tY=Y)$, hence $R_{\wstar}(f)=\E[\ell(f(X),Y)\mid\tY=Y]$. $\square$

\paragraph{Retention.}
Let $\nu(x,y,k)=\Prob(\tY=y\mid X=x,Y=y,K=k)$ be the probability that a label is kept and $\bar\nu=\E[\nu]$.
Then $\Rcor(f)=\E[\nu\,\ell(f(X),Y)]/\bar\nu$, and for a loss with values in $[0,B]$
\begin{equation}
  |\Rcor(f)-R_{\mathrm{clean}}(f)|\le\frac{B}{2}\cdot\frac{\E|\nu-\bar\nu|}{\bar\nu}.
\end{equation}
\emph{Proof.} $\Rcor-R_{\mathrm{clean}}=\E[(\nu/\bar\nu-1)\,\ell]$ and $\E[\nu/\bar\nu-1]=0$, so $\ell$ may be replaced by $\ell-B/2$, whose absolute value is at most $B/2$. $\square$
The two risks coincide when $\nu$ is constant.
Our synthetic generator flips 40\% of each client's labels, drawn uniformly at random within the client and independently of the image and the class, so the retention probability is the same for all samples. Real-world noise carries no such guarantee.

\paragraph{Proposition 2 (posterior form).}
\emph{Under (A1) $\tY\perp X\mid(Y,K)$, the class-conditional noise of~\cite{natarajan2013learning,patrini2017making}, and (A2) $X\perp K\mid Y$, for any full-support $m$ and $q_m(c\mid x)\propto p(x\mid c)\,m(c)$, the posterior $\Prob(Y=c\mid x,j,k)$, whenever its normalizer is positive, is proportional to}
\begin{equation}
  \frac{q_m(c\mid x)}{m(c)}\,\pi_k(c)\,T_k(j\mid c).
  \label{eq:supp-posterior}
\end{equation}

\noindent\emph{Proof.}
By Bayes' rule $\Prob(Y=c\mid x,j,k)\propto p(x\mid c,k)\,\pi_k(c)\,\Prob(\tY=j\mid x,c,k)$.
By (A2) the first factor is $p(x\mid c)$, by (A1) the third is $T_k(j\mid c)$, and $p(x\mid c)\propto q_m(c\mid x)/m(c)$ as a function of $c$, the identity behind prior-shift correction~\cite{saerens2002adjusting}. $\square$

\paragraph{Corollary 1 (odds decomposition) and the spectrum case.}
Dividing the $c=j$ term of \cref{eq:supp-read} by the sum of the others, with $L(c,x)=q(c\mid x)/m(c)$, $q_c=q(c\mid x)$, and $\alpha_c=q_c/(1-q_j)$, gives for \method's trust
\begin{equation}
  \frac{w}{1-w}=\frac{q_j}{1-q_j}\,\lambda_k,
\end{equation}
where
\begin{equation}
  \lambda_k=\frac{\hat\pi_k(j)\,\hat T_k(j\mid j)/m(j)}{\sum_{c\neq j}\alpha_c\,\hat\pi_k(c)\,\hat T_k(j\mid c)/m(c)}.
\end{equation}
This identity holds for the actual teacher and the estimated signature.
The same algebra applied to \cref{eq:supp-posterior}, with $q=q_m$ and the true signature, gives $\wstar/(1-\wstar)=\lambda_k\,q_j/(1-q_j)$.
The spectrum case below uses the true signature.
Under the noise spectrum, $T_k(j\mid j)=1-\rho$ and $T_k(j\mid c)=\rho/\ell_k$ if client $k$ flips class $c$ into $j$ and $0$ otherwise.
With $\pi_k=m$ the denominator is $(\rho/\ell_k)\,\eta_k$ with $\eta_k=\sum_{c\neq j}\alpha_c\ind\{c\to j\text{ at client }k\}$, so for $q_j<1$ and $\eta_k>0$, $\lambda_k=(1-\rho)\ell_k/(\rho\,\eta_k)$.
For a dispersed client every class flips into $j$, $\eta_k=1$, and with $C=100$, $\rho=0.4$, $\ell_k=99$ we get $\lambda_k=148.5$, so a teacher probability of $0.05$ (odds $0.0526$) becomes posterior odds $7.82$ and a trust of $0.887$.
For $\ell_k=1$, $\lambda_k=1.5/\eta_k$.

\paragraph{Lemma 1 (trust is monotone in the teacher's score).}
\emph{For fixed $\pi_k$, $T_k$, and $m$ with $\pi_k(j)\,T_k(j\mid j)>0$, and fixed alternatives $\alpha$, the trust is increasing in $q_j$, and $w\to1$ as $q_j\to1$.}
This follows from the decomposition, since $\lambda_k$ does not depend on $q_j$ and is positive under this condition. \method's estimates satisfy it, through the prior pseudo-counts and the diagonal constraint on the transition matrix (\cref{supp:protocol}).
If the teacher has trained on $(x_i,\ty_i)$, memorization drives $q_j$ up for exactly the labels it has fit, and trust inference multiplies the inflated odds by $\lambda_k$; out-of-fold scoring removes this dependence path.

\begin{proposition}[Plug-in error]
\label{prop:plugin}
For $\hw\in[0,1]$ with $b=\E[\hw]>0$, $d=\E[\wstar]>0$, and $\ell\in[0,B]$, define $R_{\hw}(f)=\E[\hw\ell(f(X),\tY)]/b$.
Then $|R_{\hw}(f)-\Rcor(f)|\le B\,\mathrm{TV}(\hw/b,\wstar/d)\le B\,\E|\hw-\wstar|/b$ for all $f$, where $\mathrm{TV}(u,v)=\tfrac12\E|u-v|$.
\end{proposition}

\noindent\emph{Proof.}
By Prop.~1, $R_{\hw}-\Rcor=\E[(\hw/b-\wstar/d)\,\ell(f(X),\tY)]$.
The weight difference has mean zero, so $\ell$ may be replaced by $\ell-B/2$, giving the first inequality.
For the second, $\tfrac12\E|\hw/b-\wstar/d|\le\tfrac12\big(\E|\hw-\wstar|/b+|d-b|/b\big)\le\E|\hw-\wstar|/b$. $\square$

\paragraph{Lemma 2 (stability of trust inference).}
\emph{For nonnegative $\hat U,U^\star\in\mathbb{R}^C$ with positive sums and any $j$,}
\begin{equation}
  \Big|\frac{\hat U_j}{\sum_c\hat U_c}-\frac{U^\star_j}{\sum_c U^\star_c}\Big|\le\min\Big\{1,\frac{\|\hat U-U^\star\|_1}{\sum_c U^\star_c}\Big\}.
  \label{eq:stability}
\end{equation}

\noindent\emph{Proof.}
Let $s=\hat U_j/\sum_c\hat U_c\in[0,1]$. The difference equals $\big[(1-s)(\hat U_j-U^\star_j)-s\sum_{c\neq j}(\hat U_c-U^\star_c)\big]/\sum_c U^\star_c$, whose absolute value is at most $\|\hat U-U^\star\|_1/\sum_c U^\star_c$. Both ratios lie in $[0,1]$. $\square$

With $\hat U_c=L(c,x)\,\hat\pi_k(c)\,\hat T_k(j\mid c)$ and $U^\star_c=\frac{q_m(c\mid x)}{m(c)}\,\pi_k(c)\,T_k(j\mid c)$, the two ratios are the estimated trust $w$ (\cref{eq:supp-read}) and the ideal trust $\wstar$ (Prop.~2).
Applying \cref{eq:stability} to these quantities bounds $|w-\wstar|$ by $\min\{1,\|\hat U-U^\star\|_1/\sum_c U^\star_c\}$.
It allows zeros in $T_k$.
Because $q_m(c\mid x)/m(c)\propto p(x\mid c)$ for every full-support $m$, $U^\star$ may use the same $m$ as \method, and then
\begin{multline}
  \hat U_c-U^\star_c=\frac{q(c\mid x)-q_m(c\mid x)}{m(c)}\,\hat\pi_k(c)\hat T_k(j\mid c)\\
  +\frac{q_m(c\mid x)}{m(c)}\big(\hat\pi_k(c)\hat T_k(j\mid c)-\pi_k(c)T_k(j\mid c)\big),
\end{multline}
so the teacher's error and the signature's error enter separately.
Since $|w-\wstar|\le1$ always, \cref{prop:plugin} gives $\sup_f|R_w(f)-\Rcor(f)|\le B\,\E\big[\min\{1,\|\hat U-U^\star\|_1/\sum_cU^\star_c\}\big]/\E[w]$.
The bound describes how given errors reach the trust. It does not measure how far a trained teacher is from $q_m$.

\paragraph{A reference teacher fitted to the noisy labels.}
Under (A1) and (A2), the population minimizer of the cross-entropy on the pooled noisy labels is $\bar q(u\mid x)=\Prob(\tY=u\mid X=x)=\sum_c p_g(c\mid x)\,\bar T(u\mid c)$, where $p_g$ is the clean posterior of the pooled data, with class prior $\pi_g$, and $\bar T(u\mid c)=\sum_k\Prob(K=k\mid Y=c)\,T_k(u\mid c)$ is the pooled transition matrix.
Its mean prediction is $\bar m(u)=\Prob(\tY=u)$, and with $r_c(x)=p(x\mid c)/p_g(x)$, Bayes' rule gives
\begin{equation}
  \frac{\bar q(u\mid x)}{\bar m(u)}=\sum_c\Prob(Y=c\mid\tY=u)\,r_c(x).
\end{equation}
The evidence of such a teacher for class $u$ mixes the likelihood ratios of the classes from which pooled labels $u$ come.
Let $w_{\mathrm{noisy}}$ be the trust inferred from this evidence through the true signature. Then $|w-\wstar|\le|w-w_{\mathrm{noisy}}|+|w_{\mathrm{noisy}}-\wstar|$, where Lemma~2, with the reference's unnormalized posterior in place of $U^\star$, bounds the first term and the second is the bias of the reference itself.
For balanced classes and a symmetric pooled transition matrix with flip rate $\rho$, $\Prob(Y=c\mid\tY=u)=\kappa\,\ind\{c=u\}+(1-\kappa)/C$ with $\kappa=1-\rho C/(C-1)$, so the evidence becomes $\kappa\,r_u(x)+1-\kappa$.
Write $\mu_c=\pi_k(c)T_k(j\mid c)$, $Z_1=\sum_c r_c(x)\,\mu_c$, $Z_0=\sum_c \mu_c$, and $v_k(j)=\mu_j/Z_0=\Prob(Y=j\mid\tY=j,K=k)$ for the trust that uses the client's statistics alone. For $\kappa>0$,
\begin{equation}
  w_{\mathrm{noisy}}=\omega\,\wstar+(1-\omega)\,v_k(j),
\end{equation}
where $\omega=\kappa Z_1/\big(\kappa Z_1+(1-\kappa)Z_0\big)$, and therefore $|w_{\mathrm{noisy}}-\wstar|=(1-\omega)\,|v_k(j)-\wstar|$.
Through the client's signature, even a teacher fitted to symmetric noise moves the client-only trust toward $\wstar$, and $\omega$ grows with $Z_1/Z_0$, the likelihood ratio of the image averaged over the classes from which this client writes $j$.
The pooled transition matrices of our benchmarks are not exactly symmetric (\cref{supp:benchmarks}), and shared targets (\cref{supp:shared}) concentrate $\Prob(Y=c\mid\tY=u)$ on fewer classes.

\paragraph{Trust from the student's own predictions.}
Fix a transition matrix $T$ and compute the posterior from the student, $u_c\propto p_{\theta,c}\,T(j\mid c)$, held fixed for one step.
The soft cross-entropy $-\sum_c u_c\log p_{\theta,c}$ then has logit gradient $p_\theta-u$, exactly the gradient of the forward-corrected loss $-\log\sum_c p_{\theta,c}\,T(j\mid c)$~\cite{patrini2017making}.
Training on this soft target is therefore forward correction with the student as the predictor, and it inherits whatever the student memorizes.
\method's trust instead comes from teachers never trained on the label and is frozen before the student starts.

\section{Algorithm and protocol}
\label{supp:protocol}

\begin{figure}[h]
\fbox{\parbox{0.96\linewidth}{\small
\textbf{Algorithm 1: \method.}\\[2pt]
\emph{Input:} clients $k=1..M$ with $D_k$; rounds $R_T$ (teachers), $R_S$ (student).
\begin{enumerate}[leftmargin=*,itemsep=0pt,topsep=2pt]
  \item Each client splits $D_k$ into folds $A_k,B_k$, with duplicate images and every patient they belong to in one fold, and holds out about 10\% of each fold for calibration.
  \item Train teacher A with FedAvg on $\{A_k\}$ and teacher B on $\{B_k\}$ for $R_T$ rounds (calibration samples excluded).
  \item For each teacher: fit $\tau$ from per-client sums of $\sum\log\sum_c q_\tau(c\mid x)T_0(\ty\mid c)$ on its calibration samples; compute $m$ from per-client mean calibrated predictions.
  \item Each client scores every sample with the teacher of the other fold: $L(c,x)=q(c\mid x)/m(c)$.
  \item Each client, within each outer fold and for each of three inner folds: fit $\hat\pi_k,\hat T_k$ by MAP-EM on the other two inner folds, selecting $\beta$ by held-out noisy-label likelihood; compute $w=\hat p_k(\ty\mid x,\ty)$ for the held-out inner fold.
  \item Clients send $\sum_i w_i$ and $n_k$; the server returns $\bar w$.
  \item Train the student with FedAvg for $R_S$ rounds on all original labels with loss $(w/\bar w)\,\ell_{\mathrm{SCE}}$.
\end{enumerate}}}
\end{figure}

\paragraph{Cross-fitting.}
Duplicate images, identified by fingerprint, are kept in one fold.
On ICH every slice of a patient stays in one fold of one client, one inner fold, and one calibration set.
Each teacher trains for 60 rounds with standard cross-entropy and the federation settings of the student.
Calibration samples are scored by the teacher of the other fold, like all other samples.
For the in-sample control, the same two teachers score the fold they trained on. Calibration samples, on which neither teacher trained, keep their out-of-fold scores.

\paragraph{Calibration.}
$\log\tau$ is fitted on $[-2,3]$ by a bounded scalar optimizer, each evaluation summing one value per client; the objective uses $T_0$ with diagonal $0.5$ and off-diagonal $0.5/(C-1)$.
The reference prior averages each client's mean calibrated prediction, weighted by the number of samples the teacher trained on at that client.
The diagonal $0.5$ of $T_0$ is a fixed reference value, not an estimate of the noise rate.

\paragraph{Local estimation.}
Client $k$ fits its signature on a set $S$ of its own samples, using only the evidence $L(c,x_i)=q(c\mid x_i)/m(c)$ and the observed labels $\ty_i$.
EM starts from the uniform prior and $\hat T_k=T_0$ and alternates two steps.
The E-step evaluates \cref{eq:supp-read} at the current estimates for every $i\in S$,
\begin{equation}
  \gamma_i(c)=\frac{L(c,x_i)\,\hat\pi_k(c)\,\hat T_k(\ty_i\mid c)}{\sum_{c'}L(c',x_i)\,\hat\pi_k(c')\,\hat T_k(\ty_i\mid c')}.
\end{equation}
The M-step counts with these soft assignments and adds pseudo-counts.
The prior receives 25 pseudo-counts in total, spread uniformly over the classes,
\begin{equation}
  \hat\pi_k(c)=\frac{\sum_{i\in S}\gamma_i(c)+25/C}{|S|+25}.
\end{equation}
Each row of the transition matrix receives $\beta$ pseudo-counts of $T_0$,
\begin{gather}
  N(c,j)=\beta\,T_0(j\mid c)+\sum_{i\in S:\,\ty_i=j}\gamma_i(c),\\
  \tilde T_k(j\mid c)=\frac{N(c,j)}{\sum_{j'}N(c,j')}.
\end{gather}
If a row has $\tilde T_k(c\mid c)<0.5$, its diagonal is set to $0.5$ and its off-diagonal entries are rescaled in proportion to sum to $0.5$; otherwise $\hat T_k=\tilde T_k$.
This keeps every label more likely right than wrong at each client and rules out solutions that swap classes.
On CIFAR-100N, 25 of the 100 classes are labeled correctly less than half the time, below this floor, and \method still leads there by 2.84 points (\cref{tab:main}).
For $\beta=\infty$ the matrix stays at $T_0$ and only the prior is fitted.
EM stops when the largest change in $\hat\pi_k$ and $\hat T_k$ falls below $10^{-8}$, with up to 1,500 iterations, extended to 6,000 if needed, and all fits converged.

\paragraph{Choosing $\beta$ and scoring.}
Let $\bar p_i(c)\propto L(c,x_i)\,\hat\pi_k(c)$ be the class posterior of sample $i$ before its label is seen.
A candidate fit is judged by the log-likelihood of held-out noisy labels,
\begin{equation}
  \mathcal{J}(\beta)=\sum_{i\in V}\log\sum_{c}\bar p_i(c)\,\hat T_k(\ty_i\mid c),
\end{equation}
the probability that the fitted signature assigns to the label the client actually wrote on a validation set $V$.
Each client splits its out-of-fold samples into three inner folds.
To score inner fold $h$, the other two folds $a$ and $b$ select $\beta\in\{5,25,100,400,\infty\}$ by fitting on $a$ and validating on $b$, and vice versa, and summing the two scores.
The selected $\beta$ is then refitted on $a\cup b$, and the trust of every sample in $h$ is computed from \cref{eq:supp-read} with these estimates.
No sample's own label enters the statistics that score it.

\paragraph{Student.}
$\ell_{\mathrm{SCE}}(p,j)=0.1\,(-\log p_j)-\sum_c p_c\log\tilde e_{j,c}$ with $\tilde e_{j,c}=\max(e_{j,c},10^{-4})$, where $e_j$ is the one-hot label, and probabilities are clipped at $10^{-7}$ in the reverse term.
Away from the clipping boundaries, with $A=-\log10^{-4}$, the gradient with respect to the logits is $(w/\bar w)(0.1+A\,p_j)(p-e_j)$. Trust is a fixed coefficient from outside the student, while the SCE factor grows with the student's own confidence in the observed label.
Local SGD momentum is reset every round, and the server aggregates by \cref{eq:supp-fedavg}.
Global normalization preserves differences in clients' mean inferred trust in the federated objective; the explicit FedAvg weights are unchanged.

\paragraph{Communication.}
Besides model parameters, each client sends the value of the temperature objective at every temperature the scalar optimizer queries (per teacher), its mean calibrated prediction ($C$ numbers), and $\sum_i w_i$.
No per-sample quantity, prior, or transition matrix is sent.

\section{Experimental setup}
\label{supp:benchmarks}

\paragraph{Federation and training.}
All benchmarks use 20 clients, a Dirichlet(0.5) class split, and full participation (10 to 100 clients in \cref{supp:clients}).
Each round runs 5 local epochs of SGD with momentum 0.9 and weight decay $5\times10^{-4}$. Teachers and student reset local momentum every round, while during pretraining each client keeps its momentum across rounds.
CIFAR models are ResNet-18 trained with batch 64 and learning rate 0.01, and ICH uses ResNet-18 at $224\times224$ with batch 32 and learning rate 0.001.
The student trains for 200 (CIFAR-10), 250 (CIFAR-100), 300 (CIFAR-10N/100N), and 100 (ICH) rounds, and every baseline gets the same number of downstream rounds.
We report the mean of the last 10 rounds of this budget; accuracy on the balanced CIFAR test sets equals macro recall, and on ICH we report macro recall.
Every number in the tables is the mean over training seeds 1, 2, and 3 of the last-10-round mean, with sample standard deviations across the seeds. The clients, folds, observed labels, and pretrained encoder are fixed, so the seeds measure training randomness. New class splits and noise draws appear in \cref{supp:clients}, one for each number of clients, and new error structures in \cref{supp:shared}. All diagnostics use the same three seeds, except the single-seed study of \cref{supp:shared}. Only the examples of single images and clients (\cref{fig:examples}, \cref{fig:supp-channels,fig:supp-examples}) come from one training seed, seed 1.

\paragraph{Initialization.}
Teachers and student of \method start from the same encoder, trained with FedAvg on the clients' images by contrastive learning~\cite{simclr} without labels, for 100 rounds from random weights on CIFAR and 60 rounds from ImageNet weights on ICH. The two contrastive views are a weak (random crop and flip) and a strong (RandAugment) augmentation on CIFAR, and two random flips with brightness and contrast jitter on ICH. The InfoNCE temperature is 0.07. The supervised stages use random crops and flips on CIFAR and random flips on ICH.
Baselines follow the initialization of their original methods; \cref{tab:supp-init} reruns five of them from \method's encoder.

\paragraph{Noise spectrum.}
For client $k$, $\ell_k=\max\{1,\mathrm{round}((1-s_k)(C-1))\}$ with $s_k\sim\mathrm{Beta}(0.5,0.5)$; for every class $c$, $\ell_k$ targets are drawn without replacement from the other classes, independently across clients and classes.
A fixed corruption mask, drawn uniformly at random within each client, selects 40\% of its labels; a selected label of class $c$ takes one of client $k$'s targets for $c$ uniformly at random.
\emph{CIFAR-10 instance:} $\ell_k$ for clients 0--19 is 1, 1, 1, 5, 8, 8, 4, 1, 2, 1, 3, 1, 1, 1, 1, 9, 4, 3, 2, 2; 19,992 labels are flipped (39.98\%).
\emph{CIFAR-100 instance:} $\ell_k$ is 95, 1, 58, 32, 99, 5, 47, 66, 84, 89, 19, 26, 78, 20, 6, 99, 28, 2, 76, 99; 20,001 labels are flipped (40.00\%).
Pooled over clients, every class has flip targets in all other classes, and the largest target receives 33\% (CIFAR-10) and 8\% (CIFAR-100) of a class's flipped labels on average.
\emph{ICH:} the same generator on five classes.

\paragraph{CIFAR-N.}
We use the original CIFAR images with the official \texttt{worse\_label} (CIFAR-10N) and \texttt{noisy\_label} (CIFAR-100N fine) annotations; 40.2\% of the training labels differ from the clean labels in both.
The clients split the images with the same Dirichlet(0.5) protocol; the noise therefore follows the annotators, not the clients.

\paragraph{ICH.}
We use the preprocessed five-class benchmark released with FedNoRo~\cite{fednoro}, built from the RSNA brain CT hemorrhage challenge data~\cite{ich}, with its training and test lists of 47,578 and 20,391 CT slices.
The released lists name each slice by an image ID, without patient identifiers, so we rebuild the split at the patient level.
Patient identifiers and pixel-identical slices are merged into groups, and the groups are divided 7:3 at random once.
The result is 47,672 training slices from 4,863 patients and 20,297 test slices from 2,080 patients, with no patient, study, series, or image shared between the two sets.
Training classes range from 1,139 (epidural) to 20,069 (subdural) slices, the 20 clients hold 389 to 10,813 slices each, inputs are 224$\times$224, and we report macro recall.
Every method uses the same split, clients, folds, and observed labels.
The RSNA data are public and de-identified, and no new data were collected from human subjects.
On the original slice-level lists of FedNoRo, \method is also best, with 48.01 / 49.07 / 48.05 in macro recall, precision, and F1 against 42.57 / 42.69 / 42.23 for the strongest baseline.

\section{Baselines}
\label{supp:baselines}

\begin{figure*}[t]
  \centering
  \includegraphics[width=0.98\textwidth]{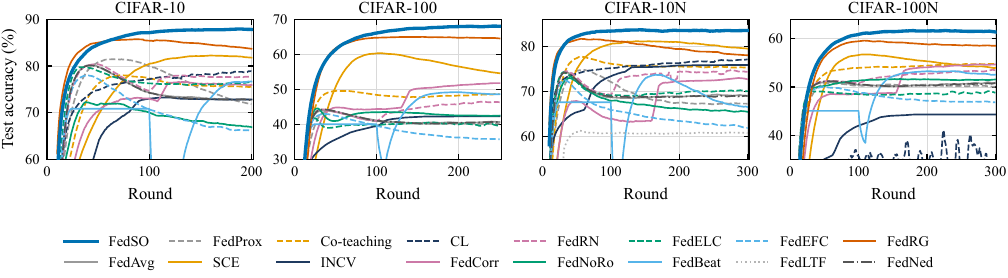}
  \caption{\textbf{Test accuracy during training} (10-round means averaged over three seeds). \method is the thick blue line. FedBeat's flat segment is its transition-estimation stage, during which the classifier is frozen.}
  \label{fig:supp-curves}
\end{figure*}

Every baseline keeps the stages and hyperparameters of its original method, adapted only to the common federation (20 clients, Dirichlet(0.5), 5 local epochs) and to the downstream budget; label-free pretraining stages (FedLTF: 100 rounds; FedRG: 100 rounds on CIFAR, 60 on ICH) are not counted in that budget.
FedBeat's second stage estimates a transition model with the classifier frozen, which shows as a flat segment in its accuracy curves.
\Cref{tab:supp-stages} lists the stage lengths of the multi-stage baselines, read from the logs of the reported runs. Stage lengths are the same in all seeds except for FedEFC.

\begin{table}[!htb]
  \centering\footnotesize
  \setlength{\tabcolsep}{2.6pt}
  \caption{\textbf{Stage lengths of the multi-stage baselines} (downstream rounds). Stages in order: FedCorr preprocessing, clean-client fine-tuning, all clients; FedNoRo and FedRN warm-up, then their main stage; FedELC warm-up, label correction; FedBeat warm-up, transition estimation, forward training; FedEFC pre-stopping (range over seeds; forward correction fills the rest of the budget); FedLTF linear probe, guided fine-tuning. FedAvg, FedProx, SCE, Co-teaching, and FedNed train in one stage, and INCV and CL screen labels before a full downstream run (see text). FedLTF and FedRG are preceded by label-free pretraining that is not counted (\cref{tab:supp-cost}). FedEFC ends pre-stopping adaptively, and on ICH it reaches forward correction in two of the three seeds, for 2 and 4 rounds.}
  \begin{tabular}{@{}lccccc@{}}
    \toprule
    Method & C10 & C100 & ICH & C10N & C100N \\
    \midrule
    FedCorr & 5/103/92 & 10/120/120 & 4/48/48 & 5/155/140 & 10/145/145 \\
    FedNoRo & 10/190 & 10/240 & 10/90 & 10/290 & 10/290 \\
    FedRN & 100/100 & 100/150 & 20/80 & 100/200 & 100/200 \\
    FedELC & 20/180 & 20/230 & 20/80 & 20/280 & 20/280 \\
    FedBeat & 15/85/100 & 15/85/150 & 10/57/33 & 15/85/200 & 15/85/200 \\
    FedEFC & 150--197 & 135--167 & 96--100 & 141--191 & 132--169 \\
    FedLTF & 20/180 & 20/230 & 20/80 & 20/280 & 20/280 \\
    FedRG & 200 & 250 & 100 & 300 & 300 \\
    \bottomrule
  \end{tabular}
  \label{tab:supp-stages}
\end{table}

\paragraph{FedBeat on ICH.}
With its default confidence threshold (0.65), FedBeat selects 553, 969, and 603 of the 47,672 slices after warm-up, 1.2\%, 2.0\%, and 1.3\%.
Its pre-registered fallback for an empty selection is therefore never used, and its 10/57/33 stages are kept.
Training on this small share of the data, the runs collapse onto different single classes: the mean per-class recall is 80.5 / 19.4 / 14.6 / 0.0 / 0.0 in seed 1, 89.2 / 0.0 / 18.3 / 0.1 / 0.0 in seed 2, and 0.0 / 0.0 / 59.5 / 40.8 / 0.0 in seed 3.

\paragraph{FedNed.}
We run the official code with execution fixes only.
FedNed needs unlabeled images on the server; we give it 128 (CIFAR-100 images for CIFAR-10/10N, CIFAR-10 images for CIFAR-100/100N, images removed from the training pool for ICH), data that no other method receives.
With its original threshold (relative entropy 0.12), its detector flags no extremely noisy client in any round, seed, or benchmark (the largest value is 0.09), so FedNed runs its FedAvg branch throughout and its mean stays within 0.1 points of FedAvg on the four CIFAR benchmarks.
In our benchmarks every client is noisy to a similar degree; what differs is the shape of the noise, which a client-level detector does not see.

\paragraph{INCV and CL.}
Both methods were proposed for centralized data. We federate them with FedAvg, so every model they train is trained across all clients and every per-sample decision is made on the client that holds the sample. All other components follow the original methods, which start from random weights.
For CL, each client splits its data into four folds by image fingerprint. Four cross-validation models are trained with FedAvg for 70 rounds, each on three folds of every client, and each client scores its fourth fold with the model that did not train on it.
Each client then applies the official cleanlab filter~\cite{confidentlearning} (pruning by both class and noise rate) to its own out-of-fold predictions, with a confident joint estimated from its own labels, and removes the flagged labels.
A fresh model is then trained with FedAvg on the kept labels, with the class weighting and loss of the original method.
INCV~\cite{incv} keeps its four screening iterations. In each, every client splits its remaining candidates into two halves, a fresh model is trained with FedAvg for 10 rounds on the selected samples and one half of every client, and each client judges its other half with this model before the halves swap.
Each client estimates its own noise rate and keeps its own selection.
The downstream stage keeps INCV's two networks with small-loss cross-updates, trained with FedAvg, and we report the first network, fixed in advance.
Neither screening stage is counted in the downstream budget.
CL keeps 53\%, 30\%, 36\%, 54\%, and 46\% of the training labels on CIFAR-10, CIFAR-100, ICH, CIFAR-10N, and CIFAR-100N, where about 60\% are correct, so on CIFAR-100 its model trains on less than a third of the data.
On the 100-class benchmarks the accuracy of CL oscillates strongly from round to round, which accounts for its large standard deviations.
On ICH, INCV concentrates on the majority class (mean per-class recall 0.0 / 20.1 / 25.0 / 26.5 / 87.9).

\paragraph{Unified initialization.}
In \cref{tab:supp-init} every method starts from the same self-supervised encoder as \method.
\method stays ahead of every baseline, and the SCE student without trust (the no-trust row of \cref{tab:ablation}) shows that the trust adds 5.70 points on CIFAR-10 and 5.62 on CIFAR-100 with the encoder held fixed.

\begin{table}[!htb]
  \centering\small
  \caption{\textbf{Unified initialization} (accuracy, \%): every method starts from the same self-supervised encoder. Mean and standard deviation over three seeds.}
  \begin{tabular}{@{}lcc@{}}
    \toprule
    Method & CIFAR-10 & CIFAR-100 \\
    \midrule
    INCV & 86.24$\pm$0.16 & 62.19$\pm$0.19 \\
    CL & 80.64$\pm$0.59 & 24.04$\pm$20.63 \\
    FedCorr & 75.02$\pm$1.26 & 56.10$\pm$0.48 \\
    FedNoRo & 73.17$\pm$0.49 & 46.04$\pm$3.86 \\
    FedELC & 74.68$\pm$1.21 & 48.98$\pm$0.76 \\
    SCE (no trust) & 82.14$\pm$0.16 & 62.41$\pm$0.15 \\
    FedRG & 83.70$\pm$0.25 & 64.55$\pm$0.08 \\
    \oursrow \method & 87.84$\pm$0.04 & 68.03$\pm$0.19 \\
    \bottomrule
  \end{tabular}
  \label{tab:supp-init}
\end{table}

\section{Full results}
\label{supp:results}

\Cref{tab:supp-sd-main,tab:supp-sd-prf} give every entry of \cref{tab:main} with its seed variation, and \cref{tab:supp-prf} lists every metric of \method.

\begin{table*}[t]
  \centering\small
  \setlength{\tabcolsep}{3pt}
  \caption{\textbf{Main results with seed variation.} Mean and sample standard deviation over training seeds 1, 2, and 3: accuracy, or macro recall on ICH (\%). \underline{Underline}: best baseline. \textit{vs.\ Best}: gain of \method over the best baseline.}
  \begin{tabular}{@{}l*{5}{>{\centering\arraybackslash}p{78pt}}@{}}
    \toprule
    Method & \multicolumn{1}{c}{CIFAR-10} & \multicolumn{1}{c}{CIFAR-100} & \multicolumn{1}{c}{ICH} & \multicolumn{1}{c}{CIFAR-10N} & \multicolumn{1}{c}{CIFAR-100N} \\
    \midrule
    FedAvg & 72.77{\scriptsize$\pm$0.53} & 40.67{\scriptsize$\pm$0.66} & 26.68{\scriptsize$\pm$0.40} & 69.11{\scriptsize$\pm$0.67} & 50.91{\scriptsize$\pm$0.05} \\
    FedProx & 71.84{\scriptsize$\pm$0.31} & 40.23{\scriptsize$\pm$0.63} & 27.08{\scriptsize$\pm$0.40} & 67.32{\scriptsize$\pm$0.49} & 50.08{\scriptsize$\pm$0.41} \\
    SCE & 81.80{\scriptsize$\pm$0.31} & 54.61{\scriptsize$\pm$0.86} & 29.52{\scriptsize$\pm$0.92} & \underline{79.50}{\scriptsize$\pm$0.12} & 53.95{\scriptsize$\pm$0.63} \\
    Co-teaching & 75.58{\scriptsize$\pm$0.25} & 48.62{\scriptsize$\pm$0.70} & 27.30{\scriptsize$\pm$0.66} & 75.31{\scriptsize$\pm$0.40} & 54.82{\scriptsize$\pm$0.11} \\
    INCV & 72.86{\scriptsize$\pm$1.15} & 42.35{\scriptsize$\pm$0.67} & 31.87{\scriptsize$\pm$1.69} & 75.91{\scriptsize$\pm$0.60} & 44.30{\scriptsize$\pm$0.51} \\
    CL & 78.88{\scriptsize$\pm$0.46} & 22.21{\scriptsize$\pm$7.07} & 28.87{\scriptsize$\pm$0.46} & 77.15{\scriptsize$\pm$0.81} & 34.21{\scriptsize$\pm$5.15} \\
    \midrule
    FedCorr & 75.95{\scriptsize$\pm$6.58} & 51.79{\scriptsize$\pm$0.63} & 23.43{\scriptsize$\pm$1.18} & 72.76{\scriptsize$\pm$2.19} & 54.67{\scriptsize$\pm$0.68} \\
    FedNoRo & 66.92{\scriptsize$\pm$2.68} & 42.52{\scriptsize$\pm$1.80} & 29.35{\scriptsize$\pm$0.45} & 65.53{\scriptsize$\pm$0.40} & 51.54{\scriptsize$\pm$0.37} \\
    FedRN & 77.71{\scriptsize$\pm$0.60} & 46.35{\scriptsize$\pm$0.28} & 28.31{\scriptsize$\pm$0.44} & 74.37{\scriptsize$\pm$0.51} & 53.46{\scriptsize$\pm$0.41} \\
    FedELC & 76.54{\scriptsize$\pm$1.83} & 39.52{\scriptsize$\pm$0.69} & 29.29{\scriptsize$\pm$0.66} & 70.16{\scriptsize$\pm$0.56} & 48.81{\scriptsize$\pm$1.03} \\
    FedBeat & 76.69{\scriptsize$\pm$0.90} & 48.64{\scriptsize$\pm$0.35} & 21.50{\scriptsize$\pm$1.43} & 66.60{\scriptsize$\pm$0.60} & 52.55{\scriptsize$\pm$0.21} \\
    FedNed & 72.76{\scriptsize$\pm$0.16} & 40.70{\scriptsize$\pm$0.60} & 27.34{\scriptsize$\pm$1.26} & 69.10{\scriptsize$\pm$0.14} & 50.83{\scriptsize$\pm$0.07} \\
    FedEFC & 66.07{\scriptsize$\pm$1.14} & 35.75{\scriptsize$\pm$0.50} & 29.65{\scriptsize$\pm$1.18} & 61.97{\scriptsize$\pm$0.02} & 46.82{\scriptsize$\pm$0.38} \\
    FedLTF & 47.53{\scriptsize$\pm$1.30} & 21.79{\scriptsize$\pm$0.92} & 30.46{\scriptsize$\pm$0.18} & 60.87{\scriptsize$\pm$2.02} & 23.43{\scriptsize$\pm$1.64} \\
    FedRG & \underline{83.70}{\scriptsize$\pm$0.25} & \underline{64.55}{\scriptsize$\pm$0.08} & \underline{43.14}{\scriptsize$\pm$0.66} & 78.06{\scriptsize$\pm$0.06} & \underline{58.55}{\scriptsize$\pm$0.43} \\
    \midrule
    \oursrow \method & \textbf{87.84}{\scriptsize$\pm$0.04} & \textbf{68.03}{\scriptsize$\pm$0.19} & \textbf{49.31}{\scriptsize$\pm$0.51} & \textbf{83.54}{\scriptsize$\pm$0.09} & \textbf{61.39}{\scriptsize$\pm$0.36} \\
    \textit{vs.\ Best} & \textit{+4.14} & \textit{+3.48} & \textit{+6.17} & \textit{+4.04} & \textit{+2.84} \\
    \bottomrule
  \end{tabular}
  \label{tab:supp-sd-main}
\end{table*}

\begin{table*}[t]
  \centering\footnotesize
  \setlength{\tabcolsep}{2pt}
  \caption{\textbf{Macro precision and macro F1 with seed variation} (mean and sample standard deviation over three training seeds, \%). Marks as in \cref{tab:supp-sd-main}.}
  \begin{tabular}{@{}lcccccccccc@{}}
    \toprule
    & \multicolumn{2}{c}{CIFAR-10} & \multicolumn{2}{c}{CIFAR-100} & \multicolumn{2}{c}{ICH} & \multicolumn{2}{c}{CIFAR-10N} & \multicolumn{2}{c}{CIFAR-100N} \\
    Method & Prec & F1 & Prec & F1 & Prec & F1 & Prec & F1 & Prec & F1 \\
    \midrule
    FedAvg & 73.27{\scriptsize$\pm$0.68} & 72.64{\scriptsize$\pm$0.52} & 40.44{\scriptsize$\pm$0.76} & 40.18{\scriptsize$\pm$0.71} & 27.39{\scriptsize$\pm$0.18} & 24.51{\scriptsize$\pm$0.20} & 69.82{\scriptsize$\pm$0.51} & 68.92{\scriptsize$\pm$0.72} & 51.79{\scriptsize$\pm$0.22} & 49.94{\scriptsize$\pm$0.05} \\
    FedProx & 72.06{\scriptsize$\pm$0.25} & 71.69{\scriptsize$\pm$0.30} & 39.70{\scriptsize$\pm$0.67} & 39.71{\scriptsize$\pm$0.64} & 27.61{\scriptsize$\pm$0.17} & 25.71{\scriptsize$\pm$0.21} & 67.60{\scriptsize$\pm$0.45} & 67.10{\scriptsize$\pm$0.45} & 50.35{\scriptsize$\pm$0.29} & 49.10{\scriptsize$\pm$0.40} \\
    SCE & 82.28{\scriptsize$\pm$0.26} & 81.63{\scriptsize$\pm$0.33} & 54.74{\scriptsize$\pm$0.92} & 54.26{\scriptsize$\pm$0.89} & 29.61{\scriptsize$\pm$0.78} & 28.45{\scriptsize$\pm$0.60} & \underline{79.95}{\scriptsize$\pm$0.10} & \underline{79.34}{\scriptsize$\pm$0.12} & 54.49{\scriptsize$\pm$0.60} & 52.80{\scriptsize$\pm$0.65} \\
    Co-teaching & 76.10{\scriptsize$\pm$0.26} & 75.39{\scriptsize$\pm$0.31} & 48.29{\scriptsize$\pm$1.20} & 48.02{\scriptsize$\pm$0.96} & 27.96{\scriptsize$\pm$0.32} & 25.84{\scriptsize$\pm$0.99} & 75.85{\scriptsize$\pm$0.34} & 75.16{\scriptsize$\pm$0.40} & 53.38{\scriptsize$\pm$0.38} & 52.97{\scriptsize$\pm$0.16} \\
    INCV & 76.28{\scriptsize$\pm$0.95} & 72.50{\scriptsize$\pm$1.37} & 41.18{\scriptsize$\pm$1.15} & 39.91{\scriptsize$\pm$0.73} & 40.12{\scriptsize$\pm$1.16} & 31.61{\scriptsize$\pm$1.59} & 77.11{\scriptsize$\pm$0.31} & 75.49{\scriptsize$\pm$0.58} & 43.59{\scriptsize$\pm$0.59} & 41.49{\scriptsize$\pm$0.42} \\
    CL & 79.78{\scriptsize$\pm$0.43} & 78.87{\scriptsize$\pm$0.48} & 32.80{\scriptsize$\pm$3.54} & 21.76{\scriptsize$\pm$6.67} & 29.38{\scriptsize$\pm$0.64} & 28.08{\scriptsize$\pm$1.11} & 77.87{\scriptsize$\pm$0.81} & 76.96{\scriptsize$\pm$0.81} & 42.65{\scriptsize$\pm$5.27} & 32.78{\scriptsize$\pm$5.19} \\
    \midrule
    FedCorr & 79.72{\scriptsize$\pm$2.91} & 74.98{\scriptsize$\pm$7.76} & 52.30{\scriptsize$\pm$0.83} & 50.73{\scriptsize$\pm$0.65} & 25.65{\scriptsize$\pm$0.22} & 18.92{\scriptsize$\pm$3.18} & 74.86{\scriptsize$\pm$1.56} & 72.41{\scriptsize$\pm$2.19} & 57.69{\scriptsize$\pm$0.75} & 53.35{\scriptsize$\pm$0.74} \\
    FedNoRo & 68.92{\scriptsize$\pm$1.81} & 66.51{\scriptsize$\pm$3.06} & 49.30{\scriptsize$\pm$8.47} & 43.41{\scriptsize$\pm$0.18} & 34.00{\scriptsize$\pm$0.61} & 28.99{\scriptsize$\pm$0.57} & 65.95{\scriptsize$\pm$0.44} & 65.24{\scriptsize$\pm$0.40} & 51.81{\scriptsize$\pm$0.34} & 50.54{\scriptsize$\pm$0.39} \\
    FedRN & 78.28{\scriptsize$\pm$0.53} & 77.51{\scriptsize$\pm$0.57} & 46.25{\scriptsize$\pm$0.13} & 45.87{\scriptsize$\pm$0.22} & 28.81{\scriptsize$\pm$0.09} & 27.30{\scriptsize$\pm$0.38} & 75.33{\scriptsize$\pm$0.66} & 74.24{\scriptsize$\pm$0.52} & 54.68{\scriptsize$\pm$0.34} & 52.52{\scriptsize$\pm$0.40} \\
    FedELC & 76.76{\scriptsize$\pm$1.78} & 76.35{\scriptsize$\pm$1.81} & 39.53{\scriptsize$\pm$1.06} & 38.51{\scriptsize$\pm$0.41} & 29.64{\scriptsize$\pm$0.42} & 27.77{\scriptsize$\pm$0.75} & 70.48{\scriptsize$\pm$0.62} & 69.97{\scriptsize$\pm$0.61} & 48.40{\scriptsize$\pm$0.83} & 47.76{\scriptsize$\pm$1.05} \\
    FedBeat & 79.50{\scriptsize$\pm$0.64} & 77.25{\scriptsize$\pm$0.81} & 46.03{\scriptsize$\pm$0.28} & 45.96{\scriptsize$\pm$0.53} & 9.55{\scriptsize$\pm$5.71} & 6.91{\scriptsize$\pm$2.54} & 68.87{\scriptsize$\pm$0.77} & 67.15{\scriptsize$\pm$0.73} & 52.72{\scriptsize$\pm$0.42} & 50.78{\scriptsize$\pm$0.13} \\
    FedNed & 73.27{\scriptsize$\pm$0.25} & 72.61{\scriptsize$\pm$0.10} & 40.51{\scriptsize$\pm$0.68} & 40.22{\scriptsize$\pm$0.63} & 27.99{\scriptsize$\pm$0.85} & 25.42{\scriptsize$\pm$0.96} & 69.78{\scriptsize$\pm$0.11} & 68.96{\scriptsize$\pm$0.15} & 51.63{\scriptsize$\pm$0.14} & 49.80{\scriptsize$\pm$0.04} \\
    FedEFC & 66.28{\scriptsize$\pm$1.18} & 65.94{\scriptsize$\pm$1.13} & 35.26{\scriptsize$\pm$0.62} & 35.29{\scriptsize$\pm$0.58} & 29.30{\scriptsize$\pm$0.78} & 28.16{\scriptsize$\pm$0.76} & 62.44{\scriptsize$\pm$0.09} & 61.86{\scriptsize$\pm$0.07} & 46.57{\scriptsize$\pm$0.52} & 45.90{\scriptsize$\pm$0.41} \\
    FedLTF & 57.74{\scriptsize$\pm$6.27} & 45.18{\scriptsize$\pm$4.36} & 44.31{\scriptsize$\pm$1.09} & 18.88{\scriptsize$\pm$0.86} & 37.29{\scriptsize$\pm$0.46} & 30.56{\scriptsize$\pm$0.11} & 66.40{\scriptsize$\pm$0.78} & 60.47{\scriptsize$\pm$1.98} & 46.87{\scriptsize$\pm$1.88} & 21.05{\scriptsize$\pm$1.36} \\
    FedRG & \underline{84.14}{\scriptsize$\pm$0.29} & \underline{83.61}{\scriptsize$\pm$0.26} & \underline{65.05}{\scriptsize$\pm$0.18} & \underline{64.17}{\scriptsize$\pm$0.08} & \underline{42.50}{\scriptsize$\pm$0.86} & \underline{42.00}{\scriptsize$\pm$0.84} & 78.52{\scriptsize$\pm$0.03} & 77.97{\scriptsize$\pm$0.05} & \underline{58.61}{\scriptsize$\pm$0.35} & \underline{57.40}{\scriptsize$\pm$0.48} \\
    \midrule
    \oursrow \method & \textbf{87.91}{\scriptsize$\pm$0.08} & \textbf{87.74}{\scriptsize$\pm$0.05} & \textbf{68.45}{\scriptsize$\pm$0.20} & \textbf{67.67}{\scriptsize$\pm$0.19} & \textbf{48.66}{\scriptsize$\pm$0.76} & \textbf{48.54}{\scriptsize$\pm$0.82} & \textbf{83.97}{\scriptsize$\pm$0.16} & \textbf{83.36}{\scriptsize$\pm$0.09} & \textbf{62.54}{\scriptsize$\pm$0.33} & \textbf{60.04}{\scriptsize$\pm$0.38} \\
    \textit{vs.\ Best} & \textit{+3.77} & \textit{+4.13} & \textit{+3.40} & \textit{+3.50} & \textit{+6.16} & \textit{+6.54} & \textit{+4.02} & \textit{+4.02} & \textit{+3.93} & \textit{+2.64} \\
    \bottomrule
  \end{tabular}
  \label{tab:supp-sd-prf}
\end{table*}

\begin{table*}[t]
  \centering\footnotesize
  \setlength{\tabcolsep}{2.0pt}
  \caption{\textbf{All score diagnostics} on the labels scored by both teachers (three-seed means). Terms as defined in \cref{supp:background}. Wrong share: fraction of the total weight carried by wrong labels (40\% without trust). Clean mean: mean trust of correct labels. ECE: expected calibration error, 10 bins. The in-sample score alone is calibrated on average but ranks labels poorly, and inferring trust from it through the signature makes it confidently wrong.}
  \begin{tabular}{@{}lcccccccccccc@{}}
    \toprule
    & \multicolumn{6}{c}{CIFAR-10} & \multicolumn{6}{c}{CIFAR-100} \\
    \cmidrule(lr){2-7}\cmidrule(l){8-13}
    Score & Acc$_\text{w}\!\downarrow$ & Rej$_\text{c}\!\downarrow$ & AUROC$\uparrow$ & wrong share$\downarrow$ & clean mean$\uparrow$ & ECE$\downarrow$ & Acc$_\text{w}\!\downarrow$ & Rej$_\text{c}\!\downarrow$ & AUROC$\uparrow$ & wrong share$\downarrow$ & clean mean$\uparrow$ & ECE$\downarrow$ \\
    \midrule
    In-sample, score only & 26.2 & 14.5 & .883 & 20.7 & .796 & .034 & 25.2 & 19.4 & .864 & 23.0 & .745 & .020 \\
    In-sample, signature & 62.7 & 0.2 & .951 & 29.1 & .992 & .240 & 96.7 & 0.0 & .920 & 38.5 & .998 & .374 \\
    Out-of-fold, score only & 2.8 & 44.3 & .917 & 6.7 & .548 & .248 & 0.1 & 70.9 & .934 & 1.5 & .317 & .407 \\
    Out-of-fold, shared signature & 14.8 & 15.9 & .924 & 14.2 & .804 & .043 & 12.3 & 17.9 & .924 & 14.3 & .787 & .049 \\
    \oursrow \method & 7.3 & 8.8 & .977 & 7.1 & .881 & .031 & 8.4 & 11.7 & .965 & 9.1 & .848 & .040 \\
    \bottomrule
  \end{tabular}
  \label{tab:supp-scores}
\end{table*}

\begin{table}[!htb]
  \centering\footnotesize
  \setlength{\tabcolsep}{2pt}
  \caption{\textbf{\method on every metric} (mean and standard deviation over three seeds). Precision, recall, and F1 are macro averages.}
  \begin{tabular}{@{}lcccc@{}}
    \toprule
    \method & Acc & Precision & Recall & F1 \\
    \midrule
    CIFAR-10 & 87.84$\pm$0.04 & 87.91$\pm$0.08 & 87.84$\pm$0.04 & 87.74$\pm$0.05 \\
    CIFAR-100 & 68.03$\pm$0.19 & 68.45$\pm$0.20 & 68.03$\pm$0.19 & 67.67$\pm$0.19 \\
    CIFAR-10N & 83.54$\pm$0.09 & 83.97$\pm$0.16 & 83.54$\pm$0.09 & 83.36$\pm$0.09 \\
    CIFAR-100N & 61.39$\pm$0.36 & 62.54$\pm$0.33 & 61.39$\pm$0.36 & 60.04$\pm$0.38 \\
    ICH & 60.81$\pm$1.25 & 48.66$\pm$0.76 & 49.31$\pm$0.51 & 48.54$\pm$0.82 \\
    \bottomrule
  \end{tabular}
  \label{tab:supp-prf}
\end{table}

\paragraph{Best windows.}
For each seed we take the best 10-round window within the budget, average it over the seeds, and compare it to the final window.
\method reaches 88.00 \vs 87.84 (CIFAR-10), 68.13 \vs 68.03 (CIFAR-100), 83.80 \vs 83.54 (CIFAR-10N), and 61.77 \vs 61.39 (CIFAR-100N).
FedRG has the highest best window among the baselines on all four, 85.81 \vs 83.70, 65.10 \vs 64.55, 81.76 \vs 78.06, and 59.66 \vs 58.55, ending between rounds 55 and 154.
Many baselines peak within the first 50 rounds and decay (\cref{fig:supp-curves}), \eg FedAvg 80.31 \vs 72.77 and FedEFC 78.05 \vs 66.07 on CIFAR-10.
On ICH the accuracy of \method peaks in the first 20 rounds (69.79 \vs 60.81) and that of the strongest baseline in the first 15 (64.66 \vs 52.08).
The best macro-recall windows are 53.62 for \method and 49.86 for the strongest baseline.

\section{Number of clients}
\label{supp:clients}

The main benchmarks fix 20 clients.
Here the same 50,000 CIFAR training images go to $M\in\{10,50,100\}$ clients.
For each $M$ a new Dirichlet(0.5) class split and a new noise spectrum are drawn before any training, with the rule and the rate ($\rho=0.4$) of the main benchmarks.
At a given $M$ every method sees the same clients, labels, and folds, and $M=20$ is the main benchmark itself.
A change of $M$ thus also changes the partition and the noise draw, so the comparison that matters is between methods at each $M$.

The baselines are FedRG and SCE, the two strongest baselines of \cref{tab:main} on both datasets, and FedCorr, a client-level method, all fixed before the runs.
As in the main experiments, \method starts from a self-supervised encoder, here trained on each new partition, and each baseline keeps the initialization of its original method.
\method keeps all its settings.
The only adaptation concerns a teacher's training half at a client that is smaller than one batch, which keeps its single partial batch. This affects three CIFAR-10 clients at $M=100$, whose smallest half has 40 images.

\method leads at every $M$ (\cref{tab:supp-clients} and \cref{fig:clients}), by 2.07 to 4.14 points on CIFAR-10 and 2.01 to 6.00 on CIFAR-100, and its worst seed exceeds the best seed of every baseline at every $M$.
The lead is largest at $M=100$ on CIFAR-100, the setting with the smallest clients.
The trust holds up on small clients.
At $M=100$ the smallest CIFAR-10 client has 116 images, and the trust still reaches AUROC .959 and .953 and puts 10.1\% and 10.7\% of the total weight on wrong labels, against 7.1\% and 9.1\% at $M=20$ and 40\% without trust.

\begin{table*}[t]
  \centering\small
  \setlength{\tabcolsep}{3.5pt}
  \caption{\textbf{Number of clients.} Accuracy (\%, mean and standard deviation over three seeds) with the 50,000 training images split among $M$ clients. $M=20$ is the main benchmark. \underline{Underline}: best baseline. Bottom rows: \method's trust on the labels scored by both teachers (AUROC, and the share of the total weight on wrong labels, 40\% without trust) and the size of the smallest client.}
  \begin{tabular}{@{}l*{8}{c}@{}}
    \toprule
    & \multicolumn{4}{c}{CIFAR-10} & \multicolumn{4}{c}{CIFAR-100} \\
    \cmidrule(lr){2-5}\cmidrule(l){6-9}
    Method & $M=10$ & $M=20$ & $M=50$ & $M=100$ & $M=10$ & $M=20$ & $M=50$ & $M=100$ \\
    \midrule
    FedRG & \underline{86.30}{\scriptsize$\pm$0.11} & \underline{83.70}{\scriptsize$\pm$0.25} & \underline{84.41}{\scriptsize$\pm$0.23} & \underline{81.74}{\scriptsize$\pm$0.08} & \underline{62.47}{\scriptsize$\pm$0.28} & \underline{64.55}{\scriptsize$\pm$0.08} & \underline{63.55}{\scriptsize$\pm$0.34} & 55.06{\scriptsize$\pm$0.33} \\
    SCE & 85.18{\scriptsize$\pm$0.19} & 81.80{\scriptsize$\pm$0.31} & 83.30{\scriptsize$\pm$0.74} & 80.51{\scriptsize$\pm$0.69} & 53.93{\scriptsize$\pm$0.66} & 54.61{\scriptsize$\pm$0.86} & 57.05{\scriptsize$\pm$0.69} & \underline{56.15}{\scriptsize$\pm$0.28} \\
    FedCorr & 49.20{\scriptsize$\pm$14.64} & 75.95{\scriptsize$\pm$6.58} & 78.14{\scriptsize$\pm$1.33} & 76.79{\scriptsize$\pm$0.50} & 51.99{\scriptsize$\pm$1.53} & 51.79{\scriptsize$\pm$0.63} & 49.15{\scriptsize$\pm$0.03} & 44.84{\scriptsize$\pm$0.18} \\
    \oursrow \method & \textbf{88.99}{\scriptsize$\pm$0.15} & \textbf{87.84}{\scriptsize$\pm$0.04} & \textbf{86.64}{\scriptsize$\pm$0.05} & \textbf{83.81}{\scriptsize$\pm$0.27} & \textbf{68.08}{\scriptsize$\pm$0.17} & \textbf{68.03}{\scriptsize$\pm$0.19} & \textbf{65.56}{\scriptsize$\pm$0.19} & \textbf{62.15}{\scriptsize$\pm$0.19} \\
    \textit{vs.\ Best} & \textit{+2.69} & \textit{+4.14} & \textit{+2.23} & \textit{+2.07} & \textit{+5.61} & \textit{+3.48} & \textit{+2.01} & \textit{+6.00} \\
    \midrule
    Trust AUROC & .976 & .977 & .965 & .959 & .962 & .965 & .960 & .953 \\
    Wrong share (\%) & 7.6 & 7.1 & 9.3 & 10.1 & 9.5 & 9.1 & 9.5 & 10.7 \\
    Smallest client & 1,962 & 994 & 395 & 116 & 4,509 & 2,124 & 721 & 367 \\
    \bottomrule
  \end{tabular}
  \label{tab:supp-clients}
\end{table*}

\section{Shared errors across clients}
\label{supp:shared}

In the noise spectrum every client draws its own flip targets, so the errors of different clients rarely coincide, and aggregation dilutes them in the teachers (\cref{sec:who}).
Here each class has one random ranking of the other classes, shared by all clients, and client $k$ sends the flipped labels of that class to the first $\ell_k$ classes of this ranking.
Clients with the same $\ell_k$ thus share their targets, and the targets of a concentrated client lie among those of a more dispersed one.
Each corrupted label keeps its slot, moving from a target of its client's original set (in class order) to the target of the same rank in the new set, and a client with $\ell_k=C-1$ keeps its labels.
The split, which labels are corrupted, how many flip from each class at each client, and every $\ell_k$ therefore stay as in the noise spectrum, and only where the flips go changes.
The teachers are retrained, calibration and signatures are refitted, and \method and the unweighted SCE student are trained with seed 1 (\cref{tab:supp-shared}).

Shared targets make the out-of-fold teacher agree with more wrong labels, as the dilution argument predicts.
On CIFAR-10 it predicts 11.2\% of them in seed 1, against 4.7\% under the noise spectrum, and the trust AUROC falls from .977 to .960.
Most CIFAR-100 clients are dispersed, so shared targets change their transition matrices little, and the teacher and the trust barely move.
Against the true classes the teacher stays calibrated (ECE .055 and .035).
The trust keeps its value for the student.
\method stays within 0.54 points of its noise-spectrum accuracy, while the unweighted student loses 2.05 and 0.38 points, so the gain grows to 7.29 and 6.27 points.

\begin{table}[!htb]
  \centering\footnotesize
  \setlength{\tabcolsep}{3pt}
  \caption{\textbf{Shared errors across clients} (seed 1). Spectrum: each client draws its own flip targets. Shared: all clients take their targets from one ranking per class, with the same corrupted samples and the same $\ell_k$. Teacher: share of wrong labels the out-of-fold teacher predicts, and its top-class calibration error against the true classes. Trust: on all 50,000 training labels, each scored once out of fold. Bottom: test accuracy (\%).}
  \begin{tabular}{@{}lcccc@{}}
    \toprule
    & \multicolumn{2}{c}{CIFAR-10} & \multicolumn{2}{c}{CIFAR-100} \\
    \cmidrule(lr){2-3}\cmidrule(l){4-5}
    & Spectrum & Shared & Spectrum & Shared \\
    \midrule
    Teacher, wrong predicted (\%)$\downarrow$ & 4.7 & 11.2 & 0.6 & 0.7 \\
    Teacher, ECE$\downarrow$ & .067 & .055 & .036 & .035 \\
    Trust, AUROC$\uparrow$ & .977 & .960 & .965 & .961 \\
    Trust, wrong share (\%)$\downarrow$ & 6.9 & 12.4 & 9.1 & 9.9 \\
    \midrule
    No trust (SCE) & 82.09 & 80.04 & 62.32 & 61.94 \\
    \oursrow \method & 87.87 & 87.33 & 68.21 & 68.21 \\
    Gain & +5.78 & +7.29 & +5.89 & +6.27 \\
    \bottomrule
  \end{tabular}
  \label{tab:supp-shared}
\end{table}

\section{Variants}
\label{supp:variants}

\begin{figure*}[t]
  \centering
  \includegraphics[width=0.98\textwidth]{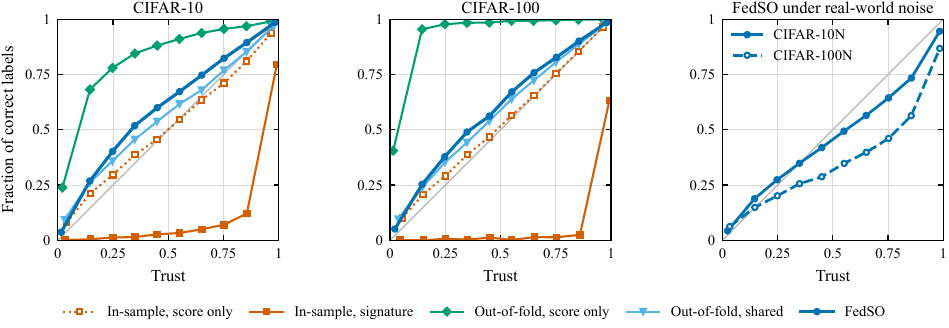}
  \caption{\textbf{Reliability of the scores} (three-seed means). Fraction of correct labels per trust bin; the diagonal is perfect calibration. Out-of-fold scores used alone lie above it (under-trust), and trust inferred from in-sample scores lies far below it (over-trust). On real-world noise (right), \method is calibrated on CIFAR-10N and over-trusts on CIFAR-100N.}
  \label{fig:supp-reliability}
\end{figure*}

\Cref{tab:supp-variants} collects the variants of the teachers and of the trust, and \cref{tab:supp-loss} crosses the student loss with the trust.
Each row reuses \method's teachers and retrains the student under the same protocol, changing only what its name indicates.
The subscript names what differs from \method.
\method$_{\text{shared}}$ takes the signature from the whole federation instead of the client. Every client infers trust from its evidence with the reference prior $m$ and one transition matrix per teacher, fitted by the same EM on the sums of all clients' expected counts, with the same inner folds and $\beta$ grid.
The other two variants change how the trust is used. \method$_{\text{mask}}$ keeps the original labels with weight $\ind\{w\ge0.5\}$, and \method$_{\text{relabel}}$ replaces labels with $w<0.5$ by the posterior's most likely class and trains with unit weights.
On the training set, the student memorizes 7.9\% and 16.1\% of the wrong labels with in-sample scores used alone, against 9.6\% and 36.3\% with trust inferred from in-sample scores, and 6.3\% and 9.2\% with \method$_{\text{shared}}$ (\method: 3.9\% and 5.5\%).
With a CE student these become 41.3\% and 88.0\% with trust inferred from in-sample scores and 11.8\% and 22.4\% with \method, and test accuracy is 75.47 and 51.92 against 84.92 and 63.81.

\begin{table}[!htb]
  \centering\small
  \newcommand{\dd}[1]{{\scriptsize\textcolor{gray}{#1}}}
  \caption{\textbf{Variants of the trust} (accuracy, \%; mean and standard deviation over three seeds). Shared: one signature for all clients, the reference prior $m$ and a jointly fitted transition matrix. Mask: weight $\ind\{w\ge0.5\}$. Relabel: labels with $w<0.5$ take the posterior's most likely class.}
  \begin{tabular}{@{}lcc@{}}
    \toprule
    Variant & CIFAR-10 & CIFAR-100 \\
    \midrule
    \oursrow \method & 87.84$\pm$0.04 & 68.03$\pm$0.19 \\
    \midrule
    In-sample, score only & 86.23$\pm$0.26 & 64.96$\pm$0.12 \\
    In-sample, signature & 85.49$\pm$0.26 & 63.30$\pm$0.16 \\
    Out-of-fold, score only & 85.78$\pm$0.04 & 61.41$\pm$0.37 \\
    \midrule
    \method$_{\text{shared}}$ & 86.62$\pm$0.20 & 66.97$\pm$0.20 \\
    \midrule
    \method$_{\text{mask}}$ & 87.32$\pm$0.08 & 67.04$\pm$0.16 \\
    \method$_{\text{relabel}}$ & 86.32$\pm$0.10 & 64.22$\pm$0.11 \\
    \midrule
    Teachers, 20 rounds & 87.96$\pm$0.02 & 68.10$\pm$0.11 \\
    Teachers, 40 rounds & 87.92$\pm$0.03 & 68.10$\pm$0.06 \\
    \bottomrule
  \end{tabular}
  \label{tab:supp-variants}
\end{table}

\begin{table}[!htb]
  \centering\small
  \caption{\textbf{Loss and trust} (mean and standard deviation over three seeds). Both help; the gains do not add up (trust adds 13.28/13.17 points to CE and 5.70/5.62 to SCE).}
  \begin{tabular}{@{}lcc@{}}
    \toprule
    Student & CIFAR-10 & CIFAR-100 \\
    \midrule
    CE, no trust & 71.64$\pm$0.51 & 50.64$\pm$0.33 \\
    SCE, no trust & 82.14$\pm$0.16 & 62.41$\pm$0.15 \\
    CE, trust & 84.92$\pm$0.22 & 63.81$\pm$0.18 \\
    \oursrow SCE, trust (\method) & 87.84$\pm$0.04 & 68.03$\pm$0.19 \\
    \bottomrule
  \end{tabular}
  \label{tab:supp-loss}
\end{table}

\paragraph{Class prior and transition matrix.}
\Cref{tab:supp-factorial} separates the two parts of the signature on the synthetic CIFAR benchmarks.
Every cell reuses \method's teachers, temperatures, and reference priors, and retrains the student under the same protocol.
A shared prior sets $\hat\pi_k=m$ at every client.
A shared transition matrix is one matrix per teacher, fitted on the sums of all clients' expected counts.
With a shared matrix and a local prior, each client keeps its own $\hat\pi_k$. With a shared prior and a local matrix, each client refits its own $\hat T_k$ with the prior held at $m$.

\method has the highest accuracy and the best ranking of labels in the table.
Against either single replacement it wins in all three seeds on both datasets.
Sharing the prior costs 0.78 and 1.06 points, and sharing the matrix costs 0.44 and 0.16.
On CIFAR-10 the two costs add up, and each stays the same whether the other part is local or shared.
On CIFAR-100 the accuracy gain of the local matrix is small (0.29, 0.18, and 0.02 points by seed) and it is absent under a shared prior (0.00).
The local matrix acts on the trust at every client. With a local prior, it lowers the share of weight on wrong labels from 9.4 to 7.1 and from 10.3 to 9.1, and after averaging over seeds it does so at all 20 clients of both datasets.
A shared matrix is also more permissive. It accepts more wrong labels and rejects slightly fewer correct ones.
Under a shared prior, the local matrix does not improve the ranking (AUROC .922 against .924 and .924 against .924).

\begin{figure}[t]
  \centering
  \includegraphics[width=\linewidth]{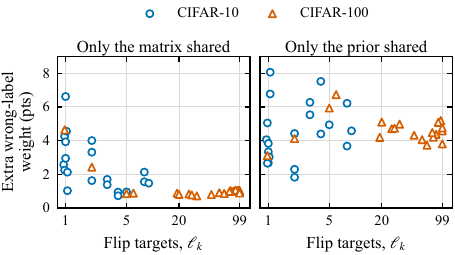}
  \caption{\textbf{The transition matrix and the prior act on different clients.} Each marker is one client (three-seed mean). Vertical: extra points of total weight on wrong labels when one part of the signature is shared and the other stays local. Sharing the matrix costs the most at $\ell_k=1$. Sharing the prior does not follow $\ell_k$.}
  \label{fig:supp-factorial}
\end{figure}

The two parts correct different errors.
Sharing the matrix alone costs the most where noise is concentrated (\cref{fig:supp-factorial}), 3.36 points of wrong-label weight at clients with $\ell_k=1$ against 1.80 at the rest on CIFAR-10, and 4.64 against 0.98 on CIFAR-100.
CIFAR-10 has nine such clients and CIFAR-100 has one, in line with the smaller gain of the local matrix on CIFAR-100.
Sharing the prior alone does not follow $\ell_k$ (4.38 against 4.70, and 3.11 against 4.69).
The prior instead moves trust toward the classes a client holds.
Take the labels that a client writes at less than half the pooled share of their class, 7.8\% and 10.0\% of all labels, of which 52.6\% and 81.0\% are wrong.
A shared prior accepts 18.9\% and 13.4\% of those wrong labels, and the local prior 6.0\% and 2.1\%.
Take the labels written at more than twice the pooled share, 50.4\% and 32.5\% of all labels.
A shared prior rejects 26.6\% and 21.9\% of the correct ones, and the local prior 7.7\% and 6.3\%.
This shift has a cost. The local prior rejects more correct rare labels (10.2\% against 7.4\%, 32.5\% against 14.3\%) and accepts more wrong common labels (9.2\% against 8.9\%, 24.5\% against 9.8\%).
The net effect on the ranking is positive (AUROC .977 and .965 against .922 and .924).
The split between the two parts depends on the class imbalance and flip patterns of the benchmark.

\begin{table*}[t]
  \centering\small
  \setlength{\tabcolsep}{4pt}
  \caption{\textbf{Class prior $\times$ transition matrix.} Accuracy (\%) with mean and standard deviation over three seeds, then three-seed means of the score diagnostics on the labels scored by both teachers, with the terms of \cref{tab:supp-scores}. Top row: \method. Bottom row: \method$_{\text{shared}}$.}
  \begin{tabular}{@{}ll*{2}{ccccc}@{}}
    \toprule
    & & \multicolumn{5}{c}{CIFAR-10} & \multicolumn{5}{c}{CIFAR-100} \\
    \cmidrule(lr){3-7}\cmidrule(l){8-12}
    Prior & Matrix & Accuracy & Acc$_\text{w}\!\downarrow$ & Rej$_\text{c}\!\downarrow$ & AUROC$\uparrow$ & wrong share$\downarrow$ & Accuracy & Acc$_\text{w}\!\downarrow$ & Rej$_\text{c}\!\downarrow$ & AUROC$\uparrow$ & wrong share$\downarrow$ \\
    \midrule
    \oursrow Local & Local & 87.84$\pm$0.04 & 7.3 & 8.8 & .977 & 7.1 & 68.03$\pm$0.19 & 8.4 & 11.7 & .965 & 9.1 \\
    Local & Shared & 87.40$\pm$0.15 & 10.6 & 7.5 & .971 & 9.4 & 67.87$\pm$0.06 & 10.1 & 11.4 & .961 & 10.3 \\
    Shared & Local & 87.06$\pm$0.16 & 10.6 & 20.9 & .922 & 11.8 & 66.97$\pm$0.14 & 11.6 & 18.9 & .924 & 13.8 \\
    Shared & Shared & 86.62$\pm$0.20 & 14.8 & 15.9 & .924 & 14.2 & 66.97$\pm$0.20 & 12.3 & 17.9 & .924 & 14.3 \\
    \bottomrule
  \end{tabular}
  \label{tab:supp-factorial}
\end{table*}

\section{Mechanism}
\label{supp:mechanism}

\begin{figure*}[t]
  \centering
  \includegraphics[width=\textwidth]{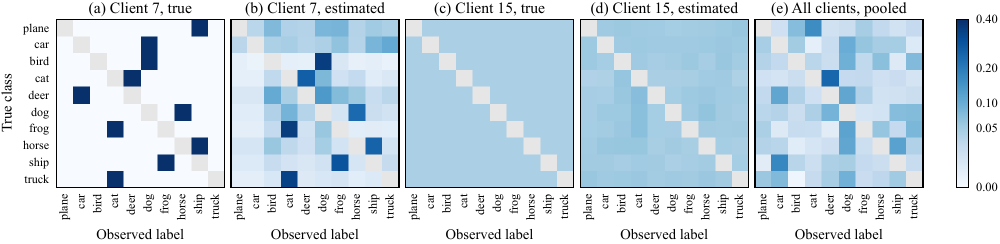}
  \caption{\textbf{Transition matrices on CIFAR-10.} Off-diagonal entries $T_k(j\mid c)$ (row: true class, column: observed label); the diagonal is shown in gray and the color scale is square-root. Client~7 ($\ell_k=1$) sends each class to a single target; \method's estimate, averaged over its six nested fits, finds these targets but also spreads some mass to other classes. Client~15 ($\ell_k=9$) flips evenly, and its estimate is nearly flat. Pooled over all 20 clients (empirical), the structure of any single client is blurred.}
  \label{fig:supp-channels}
\end{figure*}

\begin{figure}[!htb]
  \centering
  \includegraphics[width=\linewidth]{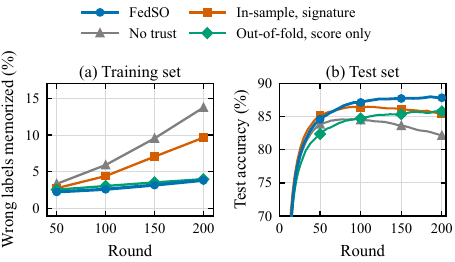}
  \caption{\textbf{What the student ends up believing on CIFAR-10} (three-seed means, rounds 50 to 200; \cref{fig:beliefs} shows CIFAR-100). (a)~Share of wrong training labels that the student memorizes. (b)~Test accuracy (10-round mean). Both failures appear, with smaller gaps.}
  \label{fig:supp-beliefs-c10}
\end{figure}

\begin{figure}[!htb]
  \centering
  \includegraphics[width=\linewidth]{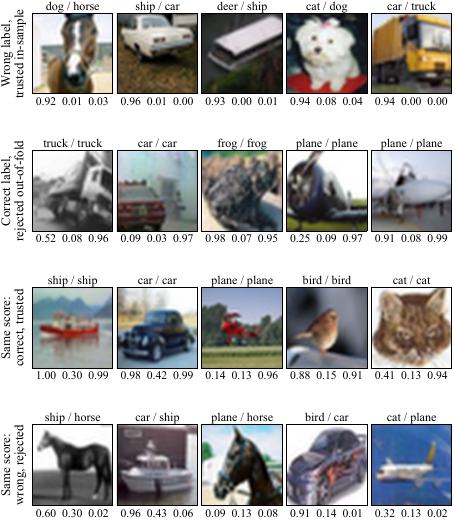}
  \caption{\textbf{Qualitative examples} (CIFAR-10). Above each image: observed label / true class. Below: in-sample and out-of-fold scores used alone, and \method trust. Images are drawn at random (fixed seed) from the labels that meet each condition, and none repeats \cref{fig:examples}. \emph{First row:} wrong labels the in-sample teacher trusts ($\ge0.9$) while the out-of-fold score and \method reject them ($<0.1$), from 504 eligible labels. \emph{Second row:} correct labels rejected by the out-of-fold score alone ($<0.1$) that \method trusts ($\ge0.9$), from 596. \emph{Last two rows:} each column is a pair with the same observed label and nearly the same out-of-fold score (within 0.005) at two different clients, one correct and trusted ($\ge0.9$) and one wrong and rejected ($\le0.1$), from 11,152 eligible pairs. The trust differs because it uses the full prediction and the client's signature.}
  \label{fig:supp-examples}
\end{figure}

\paragraph{The second opinion against the true classes.}
Over all out-of-fold predictions of the two teachers (50,000 per dataset and seed), the top-class confidence has mean ECE (10 bins) of .034 and .054 against the true classes before the temperature fit and .065 and .039 after it, against .304 and .214 for the observed labels.
The temperature fit uses noisy labels only. It sharpens the CIFAR-10 teachers slightly (fitted $\tau$ between $0.85$ and $0.96$) and leaves the CIFAR-100 teachers almost unchanged ($\tau$ between $1.01$ and $1.05$).
The prediction is thus calibrated toward the true classes, although nothing in its training targets them, and the trust that \cref{eq:supp-read} computes from the full prediction is calibrated as well (ECE .031 and .040, \cref{tab:supp-scores}).

\paragraph{Training dynamics on CIFAR-10.}
\Cref{fig:supp-beliefs-c10} repeats \cref{fig:beliefs} on CIFAR-10.

\paragraph{Estimated transition matrices and examples.}
\Cref{fig:supp-channels} shows what the local EM recovers. The flip targets of a concentrated client stand out in its estimate, while a dispersed client's estimate stays flat.
\Cref{fig:supp-examples} shows labels from both failure modes of \cref{fig:teaser} (over-trust by in-sample scores, under-trust by the out-of-fold score alone) and pairs of labels that the same out-of-fold score cannot separate but \method can.

\paragraph{The three verdicts on CIFAR-100.}
\Cref{fig:teaser} compares three verdicts on CIFAR-10, the in-sample and out-of-fold scores used alone and \method's trust, each thresholded at 0.5. Over the three seeds, on the CIFAR-100 labels scored by both teachers, the teachers predict 59.23\% (in-sample) and 0.63\% (out-of-fold) of the wrong labels, and the mean trust of correct labels is 0.317 for the out-of-fold score alone and 0.848 for \method. On CIFAR-10 the same quantities are 36.31\%, 4.86\%, 0.548, and 0.881. Grouped by the three verdicts, 16.5\% of the CIFAR-100 labels are correct and trusted by all three, 27.8\% are wrong and doubted by all, 27.2\% are correct but doubted only by the out-of-fold score, and 8.8\% are wrong but trusted only by the in-sample score (CIFAR-10: 30.8\%, 27.9\%, 17.0\%, 9.0\%).

\paragraph{Real-world noise.}
The in-sample teacher memorizes 64.19\% (CIFAR-10N) and 68.07\% (CIFAR-100N) of the wrong labels it trained on, while the out-of-fold teacher predicts 12.98\% and 12.40\%.
\method's trust has AUROC .944 and .889, and its calibration is shown in \cref{fig:supp-reliability} (right).

\paragraph{ICH by class.}
Mean per-class recall, in the order epidural, intraparenchymal, intraventricular, subarachnoid, subdural, is 5.2 / 55.7 / 66.1 / 50.6 / 69.0 for \method and 9.6 / 50.3 / 55.5 / 41.4 / 58.8 for the strongest baseline.

\section{Cost}
\label{supp:cost}

\Cref{tab:supp-cost} lists what \method adds on top of a baseline trained for the same number of student rounds.

\begin{table}[!htb]
  \centering\footnotesize
  \setlength{\tabcolsep}{4pt}
  \caption{\textbf{Cost} in full-data rounds of computation and in communication rounds. Each teacher trains on about 45\% of the data. The local estimation runs offline on each client, once before the student starts, and stores one $C$-dimensional prior and one $C\times C$ transition matrix per fit.}
  \begin{tabular}{@{}lcc@{}}
    \toprule
    Stage & Computation & Communication \\
    \midrule
    Pretraining, CIFAR / ICH & 100 / 60 & 100 / 60 \\
    Teachers, 60 rounds & $\approx$54 & 120 \\
    Teachers, 20 rounds & $\approx$18 & 40 \\
    Local estimation (offline) & on client & 0 \\
    Student & as baselines & as baselines \\
    \bottomrule
  \end{tabular}
  \label{tab:supp-cost}
\end{table}

\section{Limitations and outlook}
\label{supp:limitations}

\method's trust is as informative as its second opinion.
On CIFAR-N the errors follow the annotators and recur across clients, and those that look right to any model are partly shared by the out-of-fold teacher (12--13\%), so on CIFAR-100N \method trusts more wrong labels than under synthetic noise while still leading by 2.84 points.
For the same reason, an image-based opinion cannot tell a labeling convention from an error.
Our analysis assumes that clients share the appearance of each class and that noise within a client depends on the class only.
Natural next steps are to extend trust inference to clients whose images differ in appearance, as across hospitals, and to instance-dependent noise, and to share signatures among similar clients.

\end{document}